\documentclass{article} 
\usepackage{colm2024_conference}

\usepackage{booktabs}
\usepackage{graphicx}
\usepackage{enumitem}
\usepackage{wrapfig}
\usepackage{algorithm}
\usepackage{algpseudocode}
\usepackage{natbib}
\usepackage{makecell}
\usepackage{booktabs}
\usepackage{array}
\usepackage{amsmath}
\usepackage{amssymb}
\usepackage{amsfonts}
\usepackage{multirow}
\usepackage{verbatim}
\usepackage{caption}
\usepackage{longtable}
\usepackage{supertabular}
\usepackage{CJKutf8}
\usepackage[utf8]{inputenc} 
\usepackage[T1]{fontenc}
\usepackage[french,vietnamese,mongolian,greek,english]{babel}
\usepackage{pifont}

\usepackage{enumitem}
\usepackage{tablefootnote}
\usepackage{xspace}
\usepackage{textcomp}
\usepackage{makecell}
\usepackage{lscape} 
\usepackage{siunitx}
\usepackage{listings}
\usepackage{xcolor}
\usepackage{svg}
\definecolor{dt}{gray}{0.7}
\usepackage{pifont}       
\usepackage{bbding}       
\usepackage{fontawesome}

\usepackage{scrextend}

\usepackage{tgpagella}
\usepackage{latexsym}
\usepackage[T1]{fontenc}
\usepackage[utf8]{inputenc}
\usepackage{microtype}
\definecolor{mydarkblue}{rgb}{0,0.08,0.45}
\definecolor{citecolor}{HTML}{0071BC}
\usepackage{url}            
\usepackage{nicefrac}       
\usepackage{changepage}
\usepackage{xargs}          
\usepackage{wrapfig,lipsum,booktabs}
\usepackage{longtable}
\usepackage{subcaption}
\usepackage{endnotes}

\usepackage{pgfplots}
\usetikzlibrary{pgfplots.groupplots}
\pgfplotsset{compat=1.3}
\usepackage{tikz}
\usetikzlibrary{patterns}

\usepackage[most]{tcolorbox}
\usepackage{fancyhdr}

\usepackage{CJKutf8}

\usepackage{hyperref}
\definecolor{darkblue}{rgb}{0, 0, 0.5}
\hypersetup{colorlinks=true, citecolor=darkblue, linkcolor=darkblue, urlcolor=darkblue, bookmarksopen=false, bookmarksnumbered=true}

\usepackage[capitalize,noabbrev]{cleveref}

\crefname{section}{\S}{\S\S}
\Crefname{section}{\S}{\S\S}
\crefname{subsection}{\S\S}{\S\S}
\Crefname{subsection}{\S\S}{\S\S}
\crefformat{section}{\S#2#1#3}
\crefformat{subsection}{\S\S#2#1#3}

\crefname{table}{Table}{Tables}
\crefname{figure}{Figure}{Figures}
\crefname{algorithm}{Algorithm}{}
\crefname{equation}{eq.}{}
\crefname{appendix}{Appendix}{}
\crefformat{section}{Section #2#1#3}
\usepackage{multicol}
\usepackage{tcolorbox}
\usepackage{titlesec}
\titleformat*{\section}{\large\bfseries}
\usepackage{array} 
\newcolumntype{P}[1]{>{\centering\arraybackslash}p{#1}} 
\usepackage{multirow}

\usepackage{adjustbox}
\definecolor{objblue}{RGB}{3,139,221}  
\definecolor{attrred}{RGB}{255,67,67}    
\definecolor{easygreen}{RGB}{0,156,75}  
\definecolor{middleyellow}{RGB}{242,89,34}  
\definecolor{hardred}{RGB}{216,56,58}
\usepackage{colortbl}
\usepackage{array}

\usepackage[most]{tcolorbox}
\definecolor{BoxBackground}{RGB}{240, 240, 240} 
\definecolor{BoxFrame}{RGB}{0, 0, 0} 
\definecolor{TitleBackground}{RGB}{0, 0, 0} 
\definecolor{TitleText}{RGB}{255, 255, 255} 
\tcbset{
  academicbox/.style={
    boxsep=5pt,
    left=2pt,
    right=2pt,
    bottom=0.5pt,
    boxrule=0.5pt,
    colback=BoxBackground,
    colframe=BoxFrame,
    colbacktitle=TitleBackground,
    coltitle=TitleText,
    enhanced,
    attach boxed title to top left={yshift=-0.1in,xshift=0.1in},
    boxed title style={boxrule=0pt,colframe=white},
    title={#1},
  }
}
\newtcolorbox{AcademicBox}[1][]{academicbox=#1}

\title{FLUX-Reason-6M \& PRISM-Bench: A Million-Scale Text-to-Image Reasoning Dataset and Comprehensive Benchmark}

\author{
\bf
    Rongyao Fang\textsuperscript{1}*,\quad
    Aldrich Yu \textsuperscript{1}*,\quad
    Chengqi Duan\textsuperscript{2}*,\quad
    Linjiang Huang\textsuperscript{3},\quad
    Shuai Bai\textsuperscript{4},\\
    Yuxuan Cai\textsuperscript{4},\quad Kun Wang,\quad Si Liu\textsuperscript{3},\quad Xihui Liu\textsuperscript{2}$^\ddagger$,\quad Hongsheng Li\textsuperscript{1}$^\ddagger$ \\
    \vspace{0.5em} 
    \textsuperscript{1}CUHK \quad
    \textsuperscript{2}HKU \quad
    \textsuperscript{3}BUAA \quad \textsuperscript{4}Alibaba \\
    \vspace{0.5em}
    *Equal Contribution \quad \textsuperscript{$^\ddagger$}Corresponding Author
}

\begin{document}

\maketitle

\begin{abstract}
The advancement of open-source text-to-image (T2I) models has been hindered by the absence of large-scale, reasoning-focused datasets and comprehensive evaluation benchmarks, resulting in a performance gap compared to leading closed-source systems. To address this challenge, We introduce \textbf{FLUX-Reason-6M} and \textbf{PRISM-Bench} (\textbf{P}recise and \textbf{R}obust \textbf{I}mage \textbf{S}ynthesis \textbf{M}easurement \textbf{Bench}mark). FLUX-Reason-6M is a massive dataset consisting of \textbf{6 million} high-quality FLUX-generated images and \textbf{20 million} bilingual (English and Chinese) descriptions specifically designed to teach complex reasoning. The image are organized according to six key characteristics: \textit{Imagination}, \textit{Entity}, \textit{Text rendering}, \textit{Style}, \textit{Affection}, and \textit{Composition}, and design explicit \textbf{Generation Chain-of-Thought (GCoT)} to provide detailed breakdowns of image generation steps. The whole data curation takes \textbf{15,000 A100 GPU days}, providing the community with a resource previously unattainable outside of large industrial labs. PRISM-Bench offers a novel evaluation standard with seven distinct tracks, including a formidable \textit{Long Text} challenge using GCoT. Through carefully designed prompts, it utilizes advanced vision-language models for nuanced human-aligned assessment of prompt-image alignment and image aesthetics. Our extensive evaluation of 19 leading models on PRISM-Bench reveals critical performance gaps and highlights specific areas requiring improvement. Our dataset, benchmark, and evaluation code are released to catalyze the next wave of reasoning-oriented T2I generation.
\end{abstract}

\begin{figure}[h]
    \makebox[\linewidth]{
        \includegraphics[width=1.0\linewidth]{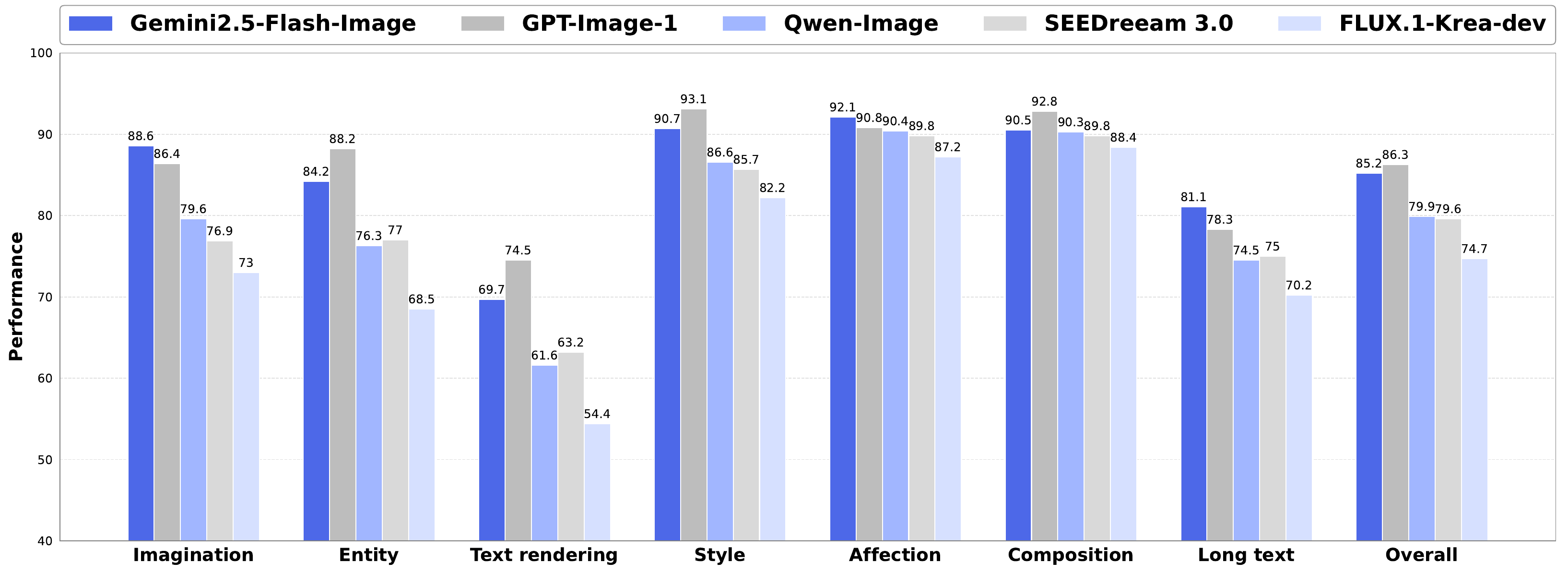}
    }
    \caption{Evaluation of state-of-the-art text-to-image models with the proposed PRISM-Bench.}
    \label{fig:teaser}
\end{figure}

\begin{figure}[p]
    \makebox[\linewidth]{
        \includegraphics[width=1.05\linewidth]{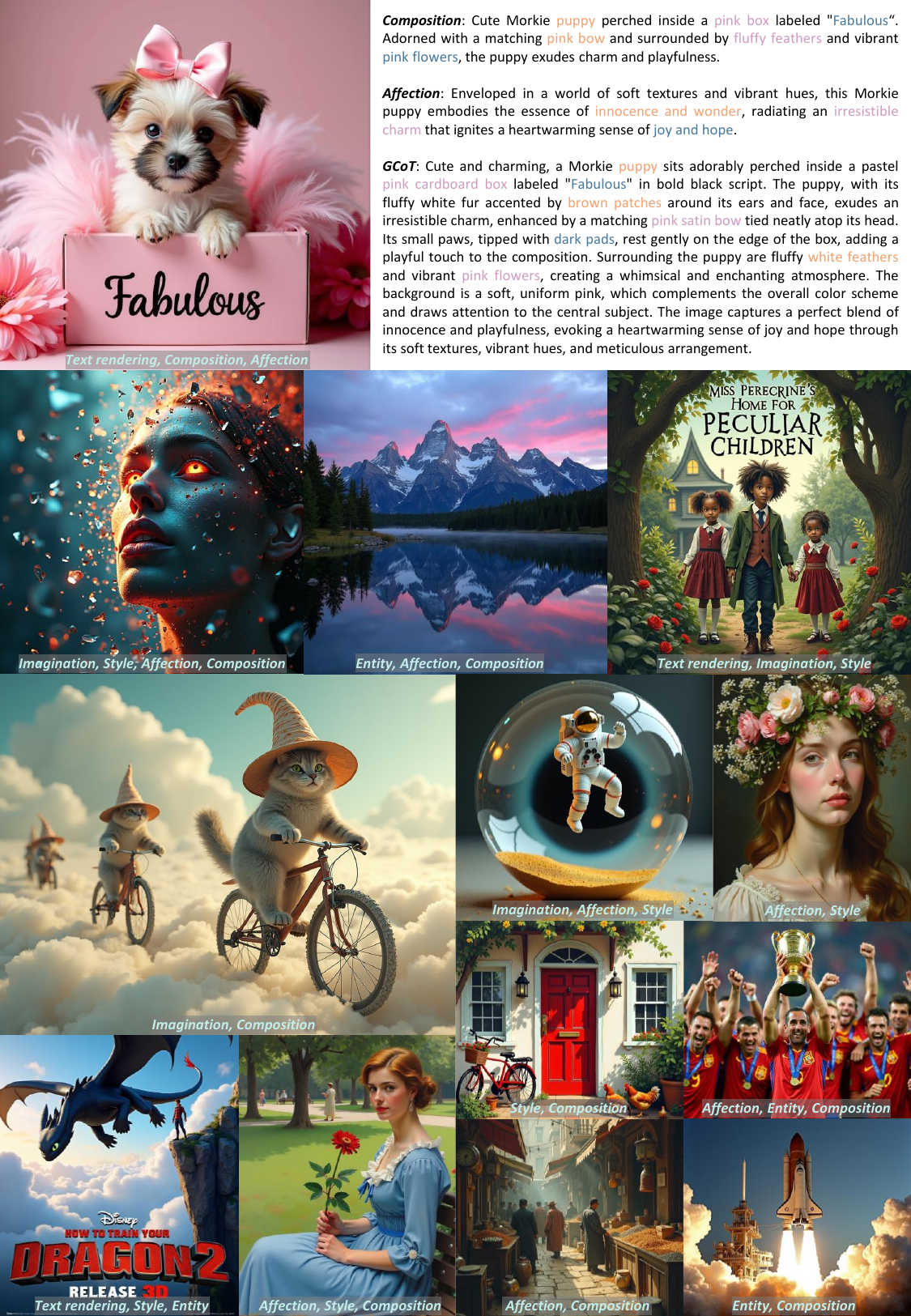}
    }
    \caption{Showcase of FLUX-Reason-6M in six different characteristics and generation chain of thought. Keywords related to characteristics in the captions are highlighted in color.}
    \label{fig:intro}
\end{figure}
 
\newpage

\section{Introduction}
Text-to-image generation models enable machines to produce engaging and coherent images, and have quickly become a key research direction in generative artificial intelligence~\citep{ho2020denoising,liu2022flow,rombach2022high,wu2022nuwa,liang2022nuwa,dalle3,podell2023sdxl,chen2023pixart,fang2024puma,duan2025got,esser2024scaling,li2024hunyuan,esser2024scaling,xie2025show,wu2025omnigen2,qin2025lumina,flux,gong2025seedream,seedream3,hidream,imagen4,gptimage,QwenImage,nanobanana}. Among these models, state-of-the-art closed-source models (e.g., Gemini2.5-Flash-Image~\citep{nanobanana}, GPT-Image-1~\citep{gptimage}) demonstrate strong instruction following and controllable synthesis capabilities, establishing new benchmarks for T2I generation. In contrast, open-source models~\citep{podell2023sdxl,chen2023pixart,SD21,SD3,SD35,zhuo2024lumina,chen2025blip3,xie2025sana} exhibit limitations when processing complex and detailed prompts.

This disparity stems from two challenges. First, the research community lacks large-scale, high-quality, and comprehensive open-source datasets. Most existing datasets consist of web-crawled image-text pairs~\citep{changpinyo2021conceptual,schuhmann2022laion,sharma2018conceptual,hu2022scaling,gadre2023datacomp}. These data are unable to be used to endow T2I models \textit{reasoning} capabilities, which is the key for synthesizing complex scenes. Although reasoning-oriented datasets exist, they tend to be narrow in scope~\citep{fang2025got}. For example, the GoT dataset~\citep{fang2025got} primarily focuses on layout planning through bounding boxes, offering limited coverage of other broader dimensions of reasoning. Second, there is an absence of a comprehensive evaluation benchmark aligned with human judgment. Most existing benchmarks~\citep{ghosh2023geneval,huang2023t2i,hu2023tifa,chefer2023attend,bakr2023hrs,feng2022training,fu2024commonsense,yu2022scaling,han2024evalmuse,hu2024ella,li2024genai,wu2024conceptmix,cho2023davidsonian} evaluate only a limited number of dimensions while neglecting key aspects such as imaginative capacity and emotional expression. Additionally, these benchmarks rely on object detectors~\citep{ghosh2023geneval} and crude CLIP scores~\citep{hessel2021clipscore,lin2024evaluating,wu2023human}, resulting in evaluation metrics that are easily saturated and fail to effectively differentiate the model's actual performance.

To resolve these problems, in this work, we introduce \textbf{FLUX-Reason-6M} and \textbf{PRISM-Bench}. FLUX-Reason-6M is a \textbf{6-million-scale} synthesized dataset designed to incorporate reasoning capabilities into the architecture of T2I generation. PRISM-Bench serves as a comprehensive and discriminative benchmark with \textbf{7 independent tracks} that closely align with human judgment.

To build FLUX-Reason-6M, we leverage the powerful capabilities of advanced image generation models and vision-language models to develop a robust data pipeline that includes large-scale data collection, synthesis, mining, annotation, filtering, and translation. We identify six key characteristics essential for T2I generation: \textit{Text rendering} and \textit{Composition}, which are common in existing researches~\citep{huang2023t2i,wang2025textatlas5m,chen2025postercraft,chen2023textdiffuser,tuo2023anytext}, and introduce \textit{Imagination}, \textit{Affection}, \textit{Entity}, and \textit{Style} to capture more nuanced and creative aspects of generation. Furthermore, we introduce \textbf{generation chain of thought (GCoT)}, which forms the core of our dataset. 
GCoT are detailed descriptions that break down the content and structure of images by comprehensively integrating the six characteristics instead of merely focusing on layout planning, providing supervision for training the reasoning capabilities of T2I models. 
As a result, FLUX-Reason-6M contains \textbf{6 million} high-quality images synthesized with FLUX.1-dev and \textbf{20 million} associated captions, each in both English and Chinese. On average, each image contains at least three annotations from different categories. 
The creation of this dataset takes \textbf{15,000 A100 GPU days}, likely positioning it as \textbf{the most expensive open-source dataset}, providing the open-source community with valuable resources to help train intelligent generative models. 
Examples from the dataset can be found in Figure~\ref{fig:intro}.

Building on the six characteristics and GCoT, we design PRISM-Bench. We first organize it into seven distinct tracks: the six categories from FLUX-Reason-6M and a uniquely challenging \textit{Long Text} track that leverages the GCoT captions to test models' complex instruction following ability. Each track contains 100 carefully selected and constructed prompts. We leverage the sophisticated cognitive judgment of advanced vision-language models (GPT-4.1~\citep{gpt4-1} and Qwen2.5-VL-72B~\citep{Qwen2.5-VL}) to evaluate prompt-image alignment and aesthetic quality, thereby providing a more reliable and human-aligned assessment of model performance. We evaluate 19 leading T2I models, including SOTA closed-source models such as Gemini2.5-Flash-Image~\citep{nanobanana}, GPT-Image-1~\citep{gptimage}, as well as top open-source models like Qwen-Image~\citep{QwenImage}. The results indicate that the gap between open-source and closed-source models is widening, but even the most advanced closed-source models still have room for improvement in certain dimensions. We present evaluation results of several models in Figure~\ref{fig:teaser}.

Our contributions represent a paradigm shift for reasoning-based T2I research:
\begin{itemize}
\item \textbf{FLUX-Reason-6M: A Landmark Dataset.} We release the first 6-million-scale T2I dataset engineered for reasoning, featuring 20 million bilingual captions and pioneering generation chain of thought prompts. This dataset is created using 128 A100 GPUs over a 4-month period, aiming to serve as the foundational dataset for the next generation of T2I models.
\item \textbf{PRISM-Bench: A New Standard for Evaluation.} We establish a comprehensive, seven-track benchmark that uses GPT-4.1 and Qwen2.5-VL-72B for nuanced and robust evaluation, offering the community a reliable tool to measure models' true capabilities.
\item \textbf{Actionable Insights from Extensive Benchmarking.} Our extensive and rigorous evaluation of leading models reveals the gaps between different models and potential areas for improvement, providing a clear roadmap for future research.
\item \textbf{Democratizing a Revolution in T2I.} We are publicly releasing the entire dataset, benchmark, and evaluation suite to lower the financial and computational barriers to entry, enabling researchers worldwide to build and test more capable generative models.
\end{itemize}

\section{FLUX-Reason-6M Dataset}
\begin{figure}[t]
    \makebox[\linewidth]{
        \includegraphics[width=1.0\linewidth]{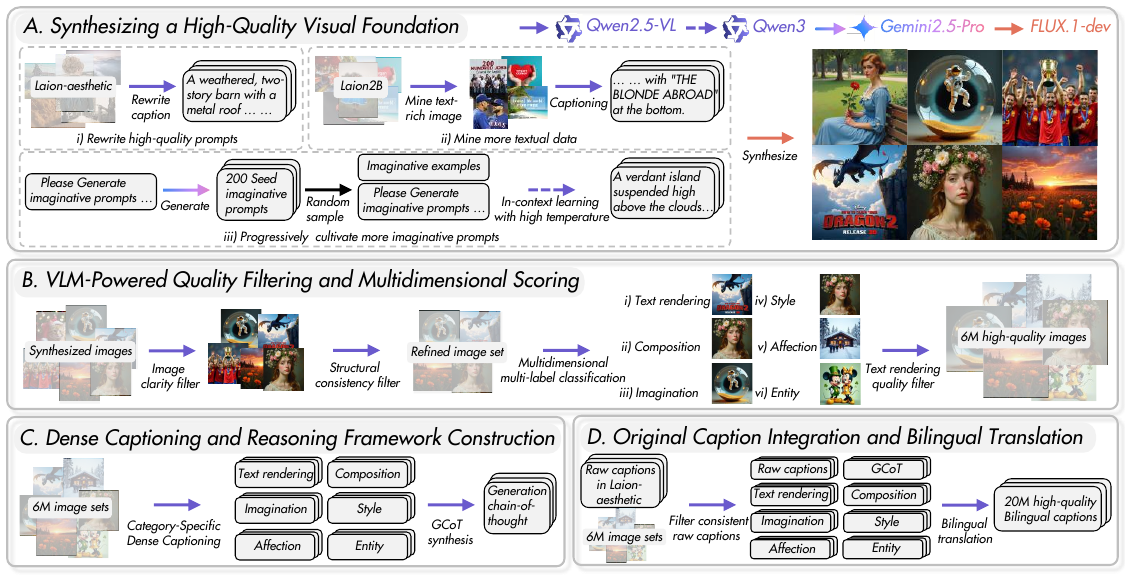}
    }
    \caption{An overview of FLUX-Reason-6M data curation pipeline. The entire process was completed using 128 A100 GPUs over a period of 4 months.}
    \label{fig:data_pipe}
\end{figure}

The central limitation of existing open-sourced T2I datasets is that they lack a structured signal for teaching models complex reasoning~\citep{podell2023sdxl,li2024playground,SD3}. They are typically flat collections of image-caption pairs that describe what is in an image but not why it is composed in a particular way. 
The recent work GoT~\citep{fang2025got} provides a 9M-sample dataset, yet this dataset is primarily assembled from existing sources (e.g., Laion-Aesthetics~\citep{schuhmann2022laion}, JourneyDB~\citep{sun2023journeydb}), leading to inconsistent quality and imbalanced distributions of image content and style. These issues stem from diverse collection and annotation protocols across source datasets.
To overcome this, we design the FLUX-Reason-6M dataset not as a mere collection of high-quality images, but as a systematic and principled framework to learn the basic rules of T2I reasoning. The overall data curation pipeline is shown in Figure~\ref{fig:data_pipe}.

\subsection{Architectural Design: The Six Characteristics and Generation Chain-of-Thought}
\paragraph{Multi-Dimensional Framework}
The essence of FLUX-Reason-6M lies in its multidimensional architectural design. We identify and define six key characteristics that are crucial for modern T2I models. These characteristics are not mutually exclusive; their intentionally designed overlapping nature aims to reflect the multifaceted aspects of complex scene synthesis, thereby providing richer and more robust training signals. The six core reasoning characteristics are:
\begin{itemize}
\item \textit{Imagination:} This category is populated with captions and images that represent surreal, fantastical, or abstract concepts. The prompts describe scenarios that defy real-world physics or combine disparate ideas in novel ways (e.g., ``a city made of glass where rivers of light flow"). The resulting images provide rich examples of creative synthesis, offering data that pushes beyond literal interpretations.
\item \textit{Entity:} This focuses on knowledge-grounded depiction. It contains image-caption pairs where the emphasis is on the accurate and detailed generation of specific real-world objects, beings, or named entities. Captions in this category are often rich with specific attributes (e.g., ``Lionel Messi dribbling past defenders in the World Cup final"), providing the model with data for high-fidelity, knowledge-aware generation.
\item \textit{Text rendering:} To address a well-known weakness in generative models, this category consists of images that successfully and legibly incorporate English text. The corresponding captions provide explicit instructions for the text's content, style, and placement within the image (e.g., ``a sign that reads `FLUX-Reason-6M' in glowing neon letters"). This provides direct and clean data for training models in typographic control.
\item \textit{Style:} This characteristic curates a vast and diverse library of artistic and photographic styles. The captions explicitly reference specific art movements (e.g., Cubism, Impressionism), visual techniques (e.g., long exposure, fisheye lens), and even the aesthetic signatures of famous artists. The images serve as high-quality examples of the successful application of these styles.
\item \textit{Affection:} This category contains image-caption pairs designed to connect abstract emotional concepts to concrete visual representations. The captions use evocative language to describe a mood, feeling, or atmosphere (e.g., ``a sense of peaceful solitude", ``a chaotic and joyful market scene"). The corresponding images translate these intangible concepts into visual cues, such as color palettes, lighting, and subject expression.
\item \textit{Composition:} This focuses on the precise arrangement and interaction of objects within a scene. The captions use explicit compositional language, including prepositions (e.g., under, behind, next to) and relative positioning. The images provide clear examples of how these complex spatial instructions are executed correctly.
\end{itemize}
A highlight of our dataset is the multi-label design. An image of ``The Eiffel Tower rendered in the style of Van Gogh's Starry Night" would be categorized under both \textit{Entity} (correctly depicting the landmark) and \textit{Style} (emulating the artist's style). This intentional overlap ensures that models learn to fuse different types of reasoning, just as a human artist would.

\paragraph{Generation Chain-of-Thought}
The cornerstone of our dataset, however, is the integration of the generation chain-of-thought. While standard captions describe image content, GCoT captions elucidate how and why the image is constructed. As illustrated in Figure~\ref{fig:intro}, this detailed step-by-step reasoning chain deconstructs the semantic and compositional logic of the final image, providing powerful intermediate supervisory signals for training. By learning these explicit reasoning paths, models can not only establish associations between vocabulary and pixels but also comprehend the underlying structures and artistic choices that constitute complex images. This structured multidimensional framework, centered on GCoT principles, forms the conceptual foundation of the entire FLUX-Reason-6M dataset.

\subsection{Synthesizing a High-Quality Visual Foundation}
Our goal is to establish a high-quality visual foundation that avoids diverse image quality in web-scraped data. Recent generative models have shown the ability to produce high-quality images. Therefore, we select the Powerful FLUX.1-dev~\citep{flux} as our synthesis engine, leveraging its advanced capabilities to generate images with exquisite detail and consistent aesthetic value. We employ a vision-language model, in conjunction with images, to rewrite captions from the Laion-Aesthetics dataset~\citep{schuhmann2022laion}, resulting in high-quality descriptions that provide a broad and versatile starting point for generation.
However, this strategy results in a biased dataset which severely lacks quantity in two characteristics: \textit{Imagination} and \textit{Text rendering}. To rectify this and ensure our dataset is balanced and comprehensive, we implement the following augmentation strategy.

\paragraph{Progressive Imagination Cultivation}
For the \textit{Imagination} category, such as uncommon scenes in daily life, we initiate a progressive generation process to produce captions of exceptional creativity and novelty. First, we leverage Gemini-2.5-Pro~\citep{gemini25} to generate a diverse set of 200 high-concept, imaginative seed prompts. In the second stage, we employ a creative expansion technique: we randomly sample 10 of these prompts and feed them into Qwen3-32B~\citep{qwen3} as in-context examples. To maximize creative output and encourage novel associations, we increase the model's temperature parameter. This process yields a vast collection of highly creative captions to push the boundaries of generative possibility. After rendering through FLUX.1-dev, these captions inject surreal and fantastical visual imagery into our dataset.

\paragraph{Mining-Generation-Synthesis Pipeline for Textual Rendering}
To address the scarcity of \textit{Text rendering} data, we develop a three-stage pipeline to harvest and regenerate high-quality textual data. First, we systematically mine Laion-2B dataset~\citep{schuhmann2022laion} using the powerful Qwen2.5-VL-32B~\citep{Qwen2.5-VL} to identify images that contain clear and legible text. Second, for each verified text-rich image, we again utilize Qwen-VL's descriptive capabilities to generate high-fidelity new captions. These captions are crafted to precisely describe the textual content, visual presentation, and contextual relationships within the image. Finally, these text-centric captions are fed into FLUX.1-dev. The final synthesis step produces images of exceptional quality where the rendered text directly corresponds to the refined caption, forming a pristine training corpus for the \textit{Text rendering} category.

This comprehensive synthesis effort, combining a high-quality baseline with targeted augmentation strategies, results in a massive pool of 8 million images. This collection provides excellent raw materials for subsequent filtering, multidimensional categorization, and dense annotation processes, ensuring that each image in the final FLUX-Reason-6M dataset meets a rigorous standard of quality and relevance.

\subsection{VLM-Powered Quality Filtering and Multidimensional Scoring}
\label{sec:score}
To transform the initial pool of 8 million synthesized images into a carefully curated resource, we design and execute a multi-stage, VLM-powered pipeline to systematically filter, categorize, and validate each image. This process ensures that all data in FLUX-Reason-6M possess both exceptional visual quality and precise categorical relevance.

\paragraph{Foundational Quality Filtering}
The first phase focuses on visual integrity. We employ Qwen-VL as an automated quality assurance inspector. Its task is to analyze each image for fundamental clarity and structural consistency. This step identifies and discards images suffering from undesirable artifacts such as excessive blurring, disruptive noise, or significant structural distortions in objects and figures. By pruning these low-quality samples, we establish a foundation of images with both aesthetic and structural integrity for the subsequent, more complex annotation and filtering phases.

\begin{figure}[t]
    \makebox[\linewidth]{
        \includegraphics[width=1.0\linewidth]{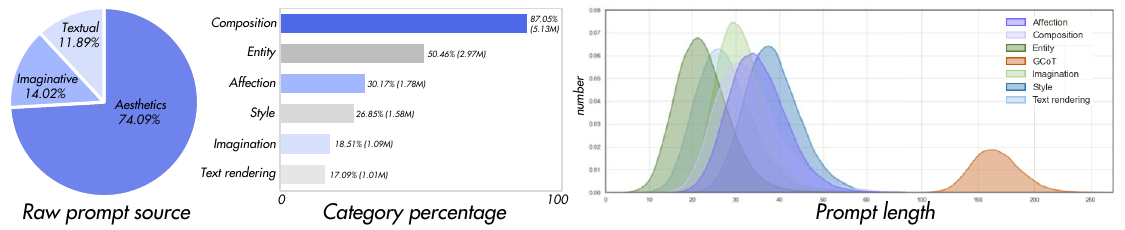}
    }
    \caption{\textbf{Left}: Three subsets of raw prompt sources. \textbf{Middle}: Image category ratio. \textbf{Right}: Prompt Suite Statistics.}
    \label{fig:data_stat}
\end{figure}

\paragraph{Robust Multidimensional Classification}
The next critical step is to organize the dataset into a multidimensional structure. We leverage Qwen-VL to evaluate each filtered image according to our six pre-defined characteristics: \textit{Imagination}, \textit{Entity}, \textit{Text rendering}, \textit{Style}, \textit{Affection}, and \textit{Composition}. Instead of simple binary classification, we adopt a quantitative scoring system that uses the model to assign a relevance score from 1 to 10 for each characteristic. After carefully setting calibrated thresholds for each characteristic, we finally determine categories for the images. This system is specifically designed for multi-labeling, capable of accurately recognizing instances where a single image belongs to multiple characteristics (such as \textit{Entity} and \textit{Style}).

\paragraph{Typographic Quality Filtering for Text Rendering}
We found that even high-quality generative models also produce illegible or contextually incorrect text.
Given the unique challenges of typographic generation, we implement a specialization filtering stage exclusively for the \textit{Text rendering} category. To ensure the dataset provides clear and reliable signals for this difficult task, we again employ Qwen-VL as a strict typographic quality inspector. It performs detailed scans of images flagged for the \textit{Text rendering} category and filters out any instances containing low-contrast, distorted, or nonsensical text. This crucial step guarantees the highest fidelity of data for this characteristic.

From the initial 8 million candidates, approximately 6 million images pass the stringent quality and relevance criteria. These images are validated for quality and tagged with rich labels that directly correspond to our six characteristics, preparing them for the final high-density annotation phase.

\subsection{VLM-Driven Dense Captioning and Reasoning Framework Construction}
With a foundation of high-quality classified images established, the next crucial phase is to generate rich, multidimensional captions and construct generation chain-of-thoughts, embedding seeds of reasoning within the dataset. This process represents a transformation from conventional captioning paradigms, moving beyond simple descriptive text to create a structured and reasoning-aware annotation framework that explicitly guides models on how to decompose and understand complex visual scenes.

\paragraph{Category-Specific Dense Captioning}
Our annotation strategy centers on leveraging the advanced multimodal reasoning capabilities of VLMs (e.g., Qwen-VL) to generate highly targeted, category-specific captions for each image. Unlike traditional approaches that produce generic descriptions~\citep{li2020oscar,jia2021scaling,zhang2021vinvl,singla2024pixels,yu2024capsfusion}, our method generates elaborate captions that emphasize the particular characteristic an image exemplifies. For instance, when processing an \textit{Entity} image, Qwen-VL is instructed to generate a caption that prioritizes accurate identification and detailed description of specific objects, landmarks, or figures present in the scene. Conversely, for images in the \textit{Style} category, the generated caption emphasizes artistic techniques, visual aesthetics, and stylistic elements that define the artistic character. This category-aware caption generation ensures that each annotation serves as a targeted training signal, teaching models to recognize and articulate the specific type required for different categories of visual content. Since each image can be assigned to multiple categories, this process ultimately forms a rich set of parallel descriptions, each providing a unique perspective for understanding the image's content and structure. The resulting annotation density far exceeds that of traditional datasets.

\paragraph{Generation Chain-of-Thought Synthesis}
The core step of our annotation process is to combine the generation chain-of-thought (GCoT), which is the main contribution and key feature of FLUX-Reason-6M. To build these reasoning processes, we employ a deliberate fusion strategy wherein Qwen-VL is provided with the full context, namely, the image together with all category-specific captions. This comprehensive input enables the model to synthesize a detailed step-by-step reasoning chain that clarifies elements present in the image but also reveals how these elements interact, why specific layouts exist, and the compositional and semantic principles governing scene composition. The resulting GCoT captions are dense, detailed narratives that serve as explicit reasoning templates. They deconstruct the image's logic layer by layer, explaining spatial relationships, artistic choices, color harmonies, emotional undertones, and compositional balance. Compared to conventional captions, these captions provide models with unprecedented insight into the creative and logical processes underlying complex image synthesis.

\subsection{Generalizable Original Caption Integration and Bilingual Release at Scale}
\paragraph{Original Caption Integration}
To broaden generalization beyond our curated reasoning signals, we reintegrate high-quality legacy captions from Laion-Aesthetics wherever they reliably describe the FLUX.1-dev synthesized images. Concretely, we use Qwen-VL as an alignment judge to score the semantic correspondence between each original Laion caption and its paired FLUX image. Captions whose scores exceed a calibrated threshold are retained as additional supervision, ensuring coverage of diverse natural language expressions while avoiding image-caption drift. After consolidation across original captions, category-specific captions, and GCoT annotations, the corpus totals \textbf{20 million} unique captions.

\paragraph{Comprehensive Bilingual Translation}
To democratize access to this powerful resource and foster international collaboration, we undertake a comprehensive translation of the entire caption corpus into Chinese. Using the advanced translation capabilities of Qwen, every original caption, category-specific caption, and GCoT caption are translated. However, for the \textit{Text rendering} category, we implement a critical content preservation strategy. To maintain the semantic integrity of the task, the specific English text intended for rendering within an image is kept in its original form in the translated caption. For example, a prompt asking for ``a sign that reads `FLUX-Reason-6M'" would be translated, but the phrase `FLUX-Reason-6M' would remain in English. This dual-language framework makes FLUX-Reason-6M one of the largest and most accessible bilingual T2I reasoning datasets, significantly broadening its impact and utility for researchers around the world.

Figure~\ref{fig:data_stat} displays the statistical characteristics of the FLUX-Reason-6M dataset, including the proportion of original prompt sources (left) and the number and percentage of each description type (middle). We also compile statistics on the word count distribution for seven categories of English descriptions and illustrate them on the right side of Figure~\ref{fig:data_stat}.

\begin{figure}[t]
    \makebox[\linewidth]{
        \includegraphics[width=1.0\linewidth]{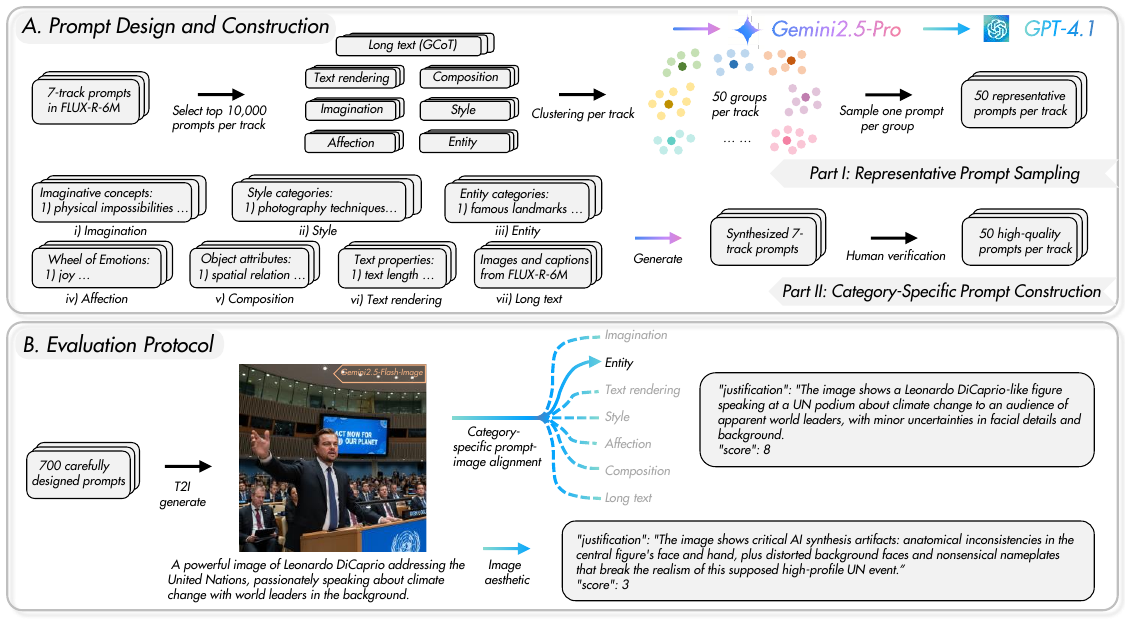}
    }
    \caption{An overview of the prompt design and evaluation protocol of PRISM-Bench.}
    \label{fig:bench_pipe}
\end{figure}

\section{PRISM-Bench}
To address the critical evaluation gap in T2I synthesis, we introduce PRISM-Bench. Existing benchmarks often lack fine granularity and cannot differentiate between state-of-the-art models, relying only on coarse metrics or narrowly defined tasks~\citep{ghosh2023geneval,huang2023t2i,hu2023tifa,chefer2023attend,bakr2023hrs,feng2022training,sun2025t2i,yu2022scaling,hu2024ella,li2024genai,wu2024conceptmix}. PRISM-Bench overcomes these limitations by providing a multidimensional, fine-grained evaluation framework. It consists of seven distinct tracks, each containing 100 carefully selected prompts designed to explore the capability boundaries of T2I models. These tracks directly correspond to the six characteristics of our dataset: \textit{Imagination}, \textit{Entity}, \textit{Style}, \textit{Text rendering}, \textit{Composition}, and \textit{Affection}, plus a challenging \textit{Long Text} constructed from our GCoT prompts. An overview of our PRISM-Bench is shown in Figure~\ref{fig:bench_pipe}.

\subsection{Prompt Design and Construction}
The 100 prompts for each track are divided into two groups of 50, designed to measure different aspects of model performance. The first group is systematically sampled from the FLUX-Reason-6M dataset to ensure broad representativeness, while the second group is carefully curated to target specific challenging aspects of each characteristic.

\paragraph{Representative Prompt Sampling}
For each track, 50 prompts come directly from the FLUX-Reason-6M dataset. To avoid selection bias and ensure broad coverage, we do not use simple random sampling but instead employ a semantic clustering and stratified sampling method. Specifically, for each category (e.g., \textit{Entity}), we collect the top 10,000 prompts from the FLUX-Reason-6M dataset that scored highest in ~\cref{sec:score}. Then we use the K-Means algorithm~\citep{krishna1999genetic} to divide the prompts into $k=50$ different semantic clusters. Each cluster represents a unique conceptual theme within that category. We select one prompt from each of the 50 groups that is closest to the group centroid (the mathematical center of the group) and remove it from the dataset. This prompt is considered the most representative example of that semantic theme. This approach guarantees conceptual diversity. It does not over-sample common themes, but instead ensures that the 50 prompts cover all concepts in the category represented in the dataset.

\paragraph{Category-Specific Prompt Construction}
The other 50 prompts for each track come from our careful curation. Specifically,
\begin{itemize}
\item \textit{Imagination:} We first divide imaginative concepts into several major categories, such as physical impossibilities and surreal narratives. Then we use an LLM (Gemini2.5-Pro) to randomly select elements from one or multiple categories to generate corresponding prompts.
\item \textit{Entity:} We curate lists of different categories of entities: famous landmarks, specific animal and plant species, historical figures, and branded objects. Then we utilize LLM to randomly select one to three entities to generate corresponding prompts.
\item \textit{Text rendering:} We design text content of varying lengths (e.g., ``FLUX-Reason-6M", ``Welcome to the future ... ..."), different font styles (e.g., handwritten script, graffiti spray paint), as well as surfaces and positions (e.g., on a wooden sign, on a t-shirt). By systematically combining elements from these three categories through LLM, we generate the corresponding prompts.
\item \textit{Style:} We define four major style categories, including art movements (e.g., Impressionism, Cubism), mediums (e.g., oil painting, watercolor), photography techniques (e.g., long exposure, macro photography), and digital/modern aesthetics (e.g., pixel art, vaporwave). These comprise a total of 25 detailed styles, and we use LLM to generate 2 prompts for each style.
\item \textit{Affection:} We use Plutchik's Wheel of Emotions~\cite{mohsin2019summarizing} as a foundational source, selecting not only the eight primary emotions (joy, trust, fear, surprise, sadness, disgust, anger, anticipation) but also their milder and more intense forms. The LLM is asked to create corresponding prompts based on these emotions.
\item \textit{Composition:} We build several attribute pools, including colors, quantities, sizes, spatial relationships, and more. For each generation, we draw several attributes from each pool and have the LLM freely combine them to create prompts featuring multiple objects with various relationships.
\item \textit{Long text:} We select 50 high-quality images from the FLUX-Reason-6M dataset along with all their corresponding captions. These are fed into Gemini2.5-Pro for long-text expansion, ultimately resulting in 50 challenging prompts.
\end{itemize}

\paragraph{PRISM-Bench-ZH}
We use Gemini2.5-Pro to translate the English prompts into Chinese, thereby creating PRISM-Bench-ZH. It is worth noting that in the \textit{Text rendering} track, we do not simply translate all text into Chinese but adapt it according to Chinese contexts. For example, ``A bottle labeled `WHISTLEPIG' featuring `SMOKED BARREL-AGED RYE' sits alongside two clear whiskey glasses, showcasing a refined presentation of the spirit" is translated as 
\begin{CJK*}{UTF8}{gbsn}
``一个标有'茅台'并写着`珍品酱香型白酒'的酒瓶，旁边放着两个透明的白酒杯，尽显这款烈酒的精致典雅。"
\end{CJK*}

\paragraph{Human-in-the-Loop Refinement}
We review all generated prompts to ensure they are unambiguous, grammatically correct, and logically sound (even if fantastical), thereby ensuring the fairness and challenging nature of the evaluation. Ultimately, we obtain 700 diverse, representative, challenging, and bilingual prompts.

\subsection{Evaluation Protocol}
To ensure a robust and nuanced evaluation of model capabilities, we develop a comprehensive assessment procedure. Our method centers on leveraging VLM's advanced cognitive abilities as a proxy for human judgment, enabling detailed analysis of model performance along two key axes: prompt-image alignment and image aesthetics. Through carefully-designed prompts, we guide VLM to evaluate generated outputs from different perspectives. This dual-metric approach provides a holistic view of each model's strengths and weaknesses. We employ GPT-4.1 and Qwen2.5-VL-72B as representative closed- and open-source VLMs, respectively, for our evaluation.

\paragraph{Fine-Grained Alignment Evaluation}
The core innovation of our approach is using track-specific evaluation prompts to assess alignment. We recognize that a generic "Does the image match the prompt?" query is insufficient to capture the specific challenges of each category, so we design customized instructions for VLM focus on the emphasis of each of the seven tracks. This ensures the evaluation addresses not just general correspondence but the success or failure of the specific task being tested by the prompt. For each generated image, VLM provides a one-sentence justification and a score from 1 (extremely poor alignment) to 10 (perfect alignment) based on the following track-specific criteria:
\begin{itemize}
\item \textit{Imagination:} The evaluation focuses on whether the model successfully synthesizes the described novel or surreal concepts, rewarding creative and coherent interpretations of imaginative ideas.
\item \textit{Entity:} The alignment score is based on the accurate rendering of specific, named real-world entities, including their key defining features and context.
\item \textit{Text rendering:} The scoring criteria are strict, focusing on the legibility, spelling accuracy, and the precise positioning of specified text within the image.
\item \textit{Style:} VLM is instructed to assess the fidelity of the generated image to the explicitly requested artistic or photographic style (e.g., ``Impressionism," ``long exposure"), checking for characteristic techniques.
\item \textit{Affection:} The assessment centers on whether the image effectively conveys the specified mood, emotion, or atmosphere through visual cues like color, lighting, and subject expression.
\item \textit{Composition:} The prompt for VLM emphasizes verifying the spatial arrangement of objects, their relative positions (e.g., ``to the left of," ``behind"), color appearance, and correct object counts as dictated by the text.
\item \textit{Long text:}  For this challenging track, the evaluation measures the model's ability to incorporate a high density of details from the complex, multi-sentence GCoT prompts.
\end{itemize}
This targeted approach allows for more precise and meaningful measurement of models' abilities across each distinct category.

\paragraph{Uniform Aesthetic Evaluation}
Unlike the alignment metric, the assessment of image aesthetics employs a single, unified set of instructions for VLM across all seven tracks. This is because aesthetic quality—encompassing factors such as lighting, color harmony, detail, and overall visual appeal—is a universal property independent of specific prompt content. VLM assigns each image a one-sentence rationale and an aesthetic score ranging from 1 (extremely low quality) to 10 (professional quality). This consistent standard ensures fair comparison of the inherent visual quality of images generated by different models.

By systematically applying this protocol to images generated by leading closed-source (e.g., Gemini2.5-Flash-Image, GPT-Image-1) and open-source (e.g., Qwen-Image, FLUX.1-Krea-dev) models for English, and Chinese-capable models (e.g., SEEDream 3.0, Qwen-Image, Bagel) for PRISM-Bench-ZH, we collect comprehensive results. Each model's performance on each track is reported as the average alignment score and aesthetic score (mapped to a 0-100 range) across the corresponding 100 prompts. The average of these two metrics represents the model's composite performance on that track. The overall average score across all 7 tracks representing the model's final performance, providing a clear and actionable overview of the current state of T2I generation.

\begin{table}[!t]\centering
\caption{Quantitative results on PRISM-Bench evaluated by GPT-4.1. Ali., Aes., and Avg. denote alignment, aesthetic, and average scores, respectively. The best result is in bold and the second best result is underlined.}
\renewcommand{\arraystretch}{1.7} 
\setlength{\tabcolsep}{3pt}

\centering
\begin{adjustbox}{width=\textwidth}
\begin{tabular}{l|ccc|ccc|ccc|ccc|ccc|ccc|ccc|ccc}
\toprule
\multirow{2}{*}{\textbf{Model}}
  & \multicolumn{3}{c|}{\textbf{Imagination}}
  & \multicolumn{3}{c|}{\textbf{Entity}}
  & \multicolumn{3}{c|}{\textbf{Text rendering}}
  & \multicolumn{3}{c|}{\textbf{Style}}
  & \multicolumn{3}{c|}{\textbf{Affection}}
  & \multicolumn{3}{c|}{\textbf{Composition}}
  & \multicolumn{3}{c|}{\textbf{Long text}}
  & \multicolumn{3}{c}{\textbf{Overall}} \\

\cmidrule(lr){2-25}

  & Ali. & Aes.& Avg.
  & Ali. & Aes.& Avg.   
  & Ali. & Aes.& Avg.     
  & Ali. & Aes.& Avg.     
  & Ali. & Aes.& Avg.     
  & Ali. & Aes.& Avg.     
  & Ali. & Aes.& Avg.     
  & Ali. & Aes.& Avg.     
  \\

\midrule

SD1.5~\citep{rombach2022high} & 36.6 & 36.1 & 36.4 & 53.8 & 41.1 & 47.5 & 8.0 & 33.1 & 20.6 & 55.3 & 55.3 & 55.3 & 64.4 & 57.5 & 61.0 & 61.1 & 51.0 & 56.1 & 35.3 & 30.4 & 32.9 & 44.9 & 43.5 & 44.2 \\

SD2.1~\citep{SD21} & 47.9 & 41.2 & 44.6 & 60.9 & 46.7 & 53.8 & 11.2 & 30.6 & 20.9 & 62.7 & 58.6 & 60.7 & 66.7 & 58.5 & 62.6 & 65.7 & 53.1 & 59.4 & 40.1 & 28.2 & 34.2 & 50.7 & 45.3 & 48.0 \\

SDXL~\citep{podell2023sdxl} & 55.3 & 61.1 & 58.2 & 72.5 & 67.4 & 70.0 & 13.8 & 37.0 & 25.4 & 72.4 & 75.4 & 73.9 & 78.9 & 77.1 & 78.0 & 75.5 & 75.3 & 75.4 & 44.2 & 39.6 & 41.9 & 58.9 & 61.8 & 60.4 \\

JanusPro-7B~\citep{januspro} & 70.4 & 65.8 & 68.1 & 67.1 & 51.9 & 59.5 & 15.5 & 36.7 & 26.1 & 71.4 & 73.8 & 72.6 & 79.2 & 71.5 & 75.4 & 83.7 & 61.0 & 72.4 & 62.4 & 39.7 & 51.1 & 64.2 & 57.2 & 60.7 \\

Playground~\citep{li2024playground} & 62.3 & 70.6 & 66.5 & 72.5 & 69.1 & 70.8 & 10.4 & 37.3 & 23.9 & 77.3 & 80.9 & 79.1 & 91.8 & 83.8 & 87.8 & 77.5 & 76.5 & 77.0 & 46.7 & 41.0 & 43.9 & 62.6 & 65.6 & 64.1 \\

FLUX.1-schnell~\citep{flux} & 63.3 & 66.2 & 64.8 & 61.8 & 51.2 & 56.5 & 46.2 & 54.1 & 50.2 & 68.6 & 70.1 & 69.4 & 75.4 & 69.9 & 72.7 & 85.1 & 67.5 & 76.3 & 69.4 & 49.7 & 59.6 & 67.1 & 61.2 & 64.2 \\

Bagel~\citep{bagel} & 69.4 & 68.0 & 68.7 & 59.0 & 50.1 & 54.6 & 30.2 & 44.5 & 37.4 & 67.9 & 71.3 & 69.6 & 81.7 & 81.4 & 81.6 & 90.5 & 73.1 & 81.8 & 68.1 & 55.3 & 61.7 & 66.7 & 63.4 & 65.1 \\

Bagel-CoT~\citep{bagel} & 68.4 & 74.2 & 71.3 & 62.4 & 60.0 & 61.2 & 23.2 & 40.1 & 31.7 & 64.4 & 70.1 & 67.3 & 87.1 & 80.5 & 83.8 & 88.5 & 77.9 & 83.2 & 64.0 & 52.0 & 58.0 & 65.4 & 65.0 & 65.2 \\

SD3-Medium~\citep{SD3} & 61.0 & 65.6 & 63.3 & 64.8 & 56.3 & 60.6 & 32.8 & 53.1 & 43.0 & 74.8 & 75.6 & 75.2 & 78.7 & 80.3 & 79.5 & 85.5 & 79.1 & 82.3 & 61.5 & 46.1 & 53.8 & 65.6 & 65.2 & 65.4 \\

SD3.5-Medium~\citep{SD35} & 69.5 & 73.0 & 71.3 & 72.8 & 63.7 & 68.3 & 33.3 & 50.1 & 41.7 & 77.4 & 80.3 & 78.9 & 84.9 & 85.5 & 85.2 & 89.4 & 79.2 & 84.3 & 63.3 & 50.5 & 56.9 & 70.1 & 68.9 & 69.5 \\

HiDream-I1-Dev~\citep{hidream} & 68.2 & 69.7 & 69.0 & 72.0 & 67.0 & 69.5 & 53.4 & 64.1 & 58.8 & 68.7 & 78.6 & 73.7 & 84.2 & 83.1 & 83.7 & 87.6 & 79.8 & 83.7 & 58.1 & 47.5 & 52.8 & 70.3 & 70.0 & 70.2 \\

SD3.5-Large~\citep{SD35} & 73.3 & 71.2 & 72.3 & 76.7 & 71.9 & 74.3 & 52.0 & 65.8 & 58.9 & 77.1 & 84.2 & 80.7 & 87.1 & 85.2 & 86.2 & 87.0 & 84.7 & 85.9 & 64.3 & 51.7 & 58.0 & 73.9 & 73.5 & 73.7 \\

FLUX.1-dev~\citep{flux} & 68.1 & 74.0 & 71.1 & 70.7 & 71.2 & 71.0 & 48.1 & 64.5 & 56.3 & 72.3 & 80.5 & 76.4 & 88.3 & \textbf{91.1} & 89.7 & 89.0 & 84.6 & 86.8 & 70.6 & 58.5 & 64.6 & 72.4 & 74.9 & 73.7 \\

FLUX.1-Krea-dev~\citep{fluxkrea} & 71.5 & 73.0 & 72.3 & 69.5 & 67.5 & 68.5 & 47.5 & 61.3 & 54.4 & 80.8 & 83.5 & 82.2 & 84.0 & 90.3 & 87.2 & 90.9 & 85.8 & 88.4 & 76.2 & 64.1 & 70.2 & 74.3 & 75.1 & 74.7 \\

HiDream-I1-Full~\citep{hidream} & 74.4 & 75.6 & 75.0 & 74.4 & 72.4 & 73.4 & 58.2 & 70.4 & 64.3 & 81.4 & 84.8 & 83.1 & 90.1 & 88.8 & 89.5 & 90.1 & 85.4 & 87.8 & 63.8 & 52.0 & 57.9 & 76.1 & 75.6 & 75.9 \\

SEEDream 3.0~\citep{seedream3} & 77.3 & 76.4 & 76.9 & 80.2 & 73.8 & 77.0 & 56.1 & 70.2 & 63.2 & 83.9 & 87.4 & 85.7 & 89.3 & 90.3 & 89.8 & 93.3 & 86.3 & 89.8 & 83.2 & 66.7 & 75.0 & 80.5 & 78.7 & 79.6 \\

Qwen-Image~\citep{QwenImage} & 80.5 & 78.6 & 79.6 & 79.3 & 73.2 & 76.3 & 54.3 & 68.9 & 61.6 & 84.5 & 88.7 & 86.6 & \underline{91.6} & 89.1 & 90.4 & \underline{93.7} & 86.9 & 90.3 & \underline{83.8} & 65.1 & 74.5 & 81.1 & 78.6 & 79.9 \\

Gemini2.5-Flash-Image~\citep{nanobanana} & \textbf{92.4} & \underline{84.8} & \textbf{88.6} & \underline{87.0} & \underline{81.3} & \underline{84.2} & \underline{65.2} & \underline{74.1} & \underline{69.7} & \underline{90.5} & \underline{90.8} & \underline{90.7} & \textbf{96.0} & 88.2 & \textbf{92.1} & 92.5 & \underline{88.5} & \underline{90.5} & \textbf{85.9} & \textbf{76.2} & \textbf{81.1} & \textbf{87.1} & \underline{83.4} & \underline{85.3} \\

GPT-Image-1 [High]~\citep{gptimage} & \underline{86.2} & \textbf{86.6} & \underline{86.4} & \textbf{90.0} & \textbf{86.3} & \textbf{88.2} & \textbf{68.8} & \textbf{80.1} & \textbf{74.5} & \textbf{92.8} & \textbf{93.3} & \textbf{93.1} & 90.7 & \underline{90.9} & \underline{90.8} & \textbf{96.2} & \textbf{89.4} & \textbf{92.8} & \underline{83.8} & \underline{72.8} & \underline{78.3} & \underline{86.9} & \textbf{85.6} & \textbf{86.3} \\

\bottomrule

\end{tabular}\label{tab:gpt_en}
\end{adjustbox}
\end{table}
\section{Experiments}
We evaluate 19 advanced image generation models on the PRISM-Bench, including Gemini2.5-Flash-Image~\citep{nanobanana}, GPT-Image-1~\citep{gptimage}, Qwen-Image~\citep{QwenImage}, SEEDream 3.0~\citep{seedream3}, FLUX series~\citep{flux,fluxkrea}, HiDream series~\citep{hidream}, Stable Diffusion series~\citep{rombach2022high,podell2023sdxl,SD21,SD3,SD35}, Playground~\citep{li2024playground}, Bagel~\citep{bagel}, and JanusPro~\citep{januspro}. The comprehensive results are shown in Table~\ref{tab:gpt_en} and Table~\ref{tab:qwen_en}. Meanwhile, we evaluate several models with Chinese language capabilities on the PRISM-Bench-ZH, including GPT-Image-1, Qwen-Image, SEEDream 3.0, HiDream series, and Bagel. The evaluation results are summarized in Table~\ref{tab:gpt_zh} and Table~\ref{tab:qwen_zh}.

\subsection{Results and Analysis on PRISM-Bench}
\paragraph{Overall Performance}
As shown in Table~\ref{tab:gpt_en} and Table~\ref{tab:qwen_en}, the overall results highlight the advantages of current SOTA closed-source models. GPT-Image-1 achieves the highest total score of 86.3, closely followed by Gemini2.5-Flash-Image with 85.3. These models outperform others across nearly all evaluation tracks. Among the remaining models, a competitive tier led by Qwen-Image is emerging. Although there is still a noticeable performance gap compared to top models, these models represent a significant leap forward from the open-source community. HiDream-I1-Full and FLUX.1-Krea-dev also achieve excellent results, indicating rapid progress in the field. Evolution within model series is also evident, with SDXL showing substantial improvement over SD1.5, while the newer SD3.5-Large further narrows the gap with top-performing models. The Qwen-VL evaluation results in Table~\ref{tab:qwen_en} largely corroborate these rankings.

\paragraph{Imagination}
Gemini2.5-Flash-Image leads by a significant margin with a high score of 88.6, while GPT-Image-1 follows closely at 86.4. This indicates that leading closed-source models possess more advanced creative interpretation capabilities. Qwen-Image's performance is also impressive, but older models like SD1.5 performs poorly, frequently generating ordinary or distorted images that failed to capture the imaginative essence of the prompts.

\begin{table}[!t]\centering
\caption{Quantitative results on PRISM-Bench evaluated by Qwen2.5-VL-72B. The best result is in bold and the second best result is underlined.}
\renewcommand{\arraystretch}{1.7} 
\setlength{\tabcolsep}{3pt}

\centering
\begin{adjustbox}{width=\textwidth}
\begin{tabular}{l|ccc|ccc|ccc|ccc|ccc|ccc|ccc|ccc}
\toprule
\multirow{2}{*}{\textbf{Model}}
  & \multicolumn{3}{c|}{\textbf{Imagination}}
  & \multicolumn{3}{c|}{\textbf{Entity}}
  & \multicolumn{3}{c|}{\textbf{Text rendering}}
  & \multicolumn{3}{c|}{\textbf{Style}}
  & \multicolumn{3}{c|}{\textbf{Affection}}
  & \multicolumn{3}{c|}{\textbf{Composition}}
  & \multicolumn{3}{c|}{\textbf{Long text}}
  & \multicolumn{3}{c}{\textbf{Overall}} \\

\cmidrule(lr){2-25}

  & Ali. & Aes.& Avg.
  & Ali. & Aes.& Avg.   
  & Ali. & Aes.& Avg.     
  & Ali. & Aes.& Avg.     
  & Ali. & Aes.& Avg.     
  & Ali. & Aes.& Avg.     
  & Ali. & Aes.& Avg.     
  & Ali. & Aes.& Avg.     
  \\

\midrule

SD1.5~\citep{rombach2022high} & 40.7 & 23.7 & 32.2 & 61.2 & 52.7 & 56.9 & 11.4 & 24.1 & 17.8 & 56.7 & 61.5 & 59.1 & 66.9 & 60.7 & 63.8 & 57.5 & 53.4 & 55.4 & 47.3 & 26.8 & 37.0 & 48.8 & 43.3 & 46.0 \\

SD2.1~\citep{SD21} & 48.9 & 28.4 & 38.6 & 66.0 & 57.6 & 61.8 & 16.7 & 31.4 & 24.0 & 62.7 & 66.5 & 64.6 & 68.5 & 62.1 & 65.3 & 64.8 & 58.3 & 61.5 & 50.7 & 29.8 & 40.2 & 54.0 & 47.7 & 50.8 \\

SDXL~\citep{podell2023sdxl} & 54.5 & 34.1 & 44.3 & 71.1 & 65.0 & 68.0 & 18.6 & 37.3 & 27.9 & 71.7 & 72.6 & 72.1 & 78.7 & 66.5 & 72.6 & 72.2 & 67.8 & 70.0 & 54.1 & 34.5 & 44.3 & 60.1 & 54.0 & 57.0 \\

Playground~\citep{li2024playground} & 59.0 & 39.0 & 49.0 & 69.4 & 56.7 & 63.0 & 15.3 & 31.9 & 23.6 & 74.6 & 74.6 & 74.6 & 88.8 & 66.0 & 77.4 & 72.2 & 61.3 & 66.7 & 56.0 & 35.3 & 45.6 & 62.2 & 52.1 & 57.1 \\

Bagel~\citep{bagel} & 68.0 & 45.0 & 56.5 & 67.6 & 53.4 & 60.5 & 29.4 & 42.3 & 35.8 & 69.0 & 69.7 & 69.3 & 87.1 & 66.7 & 76.9 & 86.6 & 69.2 & 77.9 & 64.5 & 50.2 & 57.3 & 67.5 & 56.6 & 62.0 \\

Bagel-CoT~\citep{bagel} & 68.0 & 44.1 & 56.0 & 67.6 & 53.4 & 60.5 & 29.4 & 42.3 & 35.8 & 69.0 & 69.7 & 69.3 & 87.1 & 66.7 & 76.9 & 86.6 & 69.2 & 77.9 & 64.5 & 50.2 & 57.3 & 67.5 & 56.5 & 62.0 \\

JanusPro-7B~\citep{januspro} & 65.0 & 38.8 & 51.9 & 68.6 & 63.5 & 66.0 & 23.1 & 50.3 & 36.7 & 70.7 & 75.2 & 72.9 & 80.7 & 68.0 & 74.3 & 82.4 & 71.1 & 76.7 & 63.9 & 49.0 & 56.4 & 64.9 & 59.4 & 62.1 \\

FLUX.1-schnell~\citep{flux} & 62.8 & 35.6 & 49.2 & 64.8 & 56.8 & 60.8 & 54.3 & 68.1 & 61.2 & 70.3 & 71.5 & 70.9 & 75.4 & 65.9 & 70.6 & 81.7 & 75.6 & 78.6 & 68.7 & 54.4 & 61.5 & 68.3 & 61.1 & 64.7 \\

SD3-Medium~\citep{SD3} & 64.3 & 37.7 & 51.0 & 69.4 & 63.3 & 66.3 & 38.5 & 63.3 & 50.9 & 74.6 & 79.5 & 77.0 & 80.5 & 75.5 & 78.0 & 85.6 & 79.5 & 82.5 & 63.4 & 50.3 & 56.8 & 68.0 & 64.2 & 66.1 \\

SD3.5-Medium~\citep{SD35} & 65.1 & 34.7 & 49.9 & 72.5 & 70.9 & 71.7 & 36.6 & 64.5 & 50.5 & 75.5 & 80.0 & 77.7 & 81.8 & 73.9 & 77.9 & 85.4 & 81.0 & 83.2 & 63.5 & 50.6 & 57.0 & 68.6 & 65.1 & 66.8 \\

FLUX.1-dev~\citep{flux} & 65.5 & 42.9 & 54.2 & 70.6 & 61.9 & 66.2 & 52.3 & 73.0 & 62.6 & 72.6 & 74.2 & 73.4 & 86.0 & 72.9 & 79.4 & 87.4 & 75.8 & 81.6 & 70.5 & 53.8 & 62.1 & 72.1 & 64.9 & 68.5 \\

HiDream-I1-Dev~\citep{hidream} & 68.8 & \underline{45.8} & 57.3 & 73.5 & 68.1 & 70.8 & 56.7 & 75.7 & 66.2 & 70.2 & 77.4 & 73.8 & 88.2 & 74.3 & 81.2 & 84.7 & 78.5 & 81.6 & 64.0 & 49.3 & 56.6 & 72.3 & 67.0 & 69.6 \\

SD3.5-Large~\citep{SD35} & 66.7 & 43.4 & 55.0 & 76.8 & 72.7 & 74.8 & 53.6 & 73.1 & 63.3 & 77.3 & 78.2 & 77.7 & 85.6 & 73.9 & 79.7 & 87.8 & 80.9 & 84.3 & 65.8 & 52.2 & 59.0 & 73.4 & 67.8 & 70.6 \\

HiDream-I1-Full~\citep{hidream} & 73.0 & 44.0 & 58.5 & 76.3 & 72.8 & 74.5 & 60.5 & 76.4 & 68.4 & 81.4 & 81.5 & 81.4 & 90.0 & 76.6 & 83.3 & 88.5 & 80.3 & 84.4 & 66.3 & 48.6 & 57.4 & 76.6 & 68.6 & 72.6 \\

FLUX.1-Krea-dev~\citep{fluxkrea} & 69.6 & 43.1 & 56.3 & 72.2 & 70.7 & 71.4 & 51.7 & 76.1 & 63.9 & 80.0 & \underline{86.6} & 83.3 & 82.6 & \underline{78.7} & 80.6 & 90.8 & \underline{87.1} & 88.9 & 73.6 & 73.4 & 73.5 & 74.4 & 73.7 & 74.0 \\

Qwen-Image~\citep{QwenImage} & 75.5 & 37.4 & 56.5 & 79.5 & 64.5 & 72.0 & 57.9 & 71.2 & 64.5 & 86.6 & 84.4 & 85.5 & 89.9 & 70.4 & 80.1 & \textbf{93.9} & 79.5 & 86.7 & \underline{76.8} & 70.9 & 73.8 & 80.0 & 68.3 & 74.1 \\

SEEDream 3.0~\citep{seedream3} & 75.8 & 38.0 & 56.9 & 81.3 & 74.2 & 77.7 & 58.8 & 74.0 & 66.4 & 84.4 & 84.1 & 84.2 & \underline{90.5} & 74.6 & 82.5 & \underline{93.6} & 85.1 & \underline{89.3} & 76.2 & 76.4 & 76.3 & 80.1 & 72.3 & 76.2 \\

Gemini2.5-Flash-Image~\citep{nanobanana} & \textbf{84.7} & 38.1 & \underline{61.4} & \underline{86.0} & \underline{76.7} & \underline{81.3} & \textbf{72.8} & \underline{84.3} & \textbf{78.5} & \textbf{89.5} & \textbf{87.8} & \textbf{88.6} & \textbf{94.3} & 74.8 & \textbf{84.5} & 91.2 & \textbf{88.2} & \textbf{89.7} & 76.3 & \textbf{80.6} & \textbf{78.4} & \textbf{85.0} & \underline{75.8} & \underline{80.4} \\

GPT-Image-1 [High]~\citep{gptimage} & \underline{79.8} & \textbf{53.3} & \underline{66.6} & \textbf{87.3} & \textbf{81.0} & \textbf{84.1} & \underline{66.7} & \textbf{86.8} & \underline{76.8} & \underline{87.3} & \textbf{87.8} & \underline{87.5} & 88.1 & \textbf{79.8} & \underline{84.0} & 92.2 & 84.9 & 88.5 & \textbf{77.2} & \underline{77.5} & \underline{77.4} & \underline{82.7} & \textbf{78.7} & \textbf{80.7} \\

\bottomrule

\end{tabular}\label{tab:qwen_en}
\end{adjustbox}
\end{table}

\paragraph{Entity}
GPT-Image-1 excels in this domain, achieving the highest score of 88.2, demonstrating its robust internal knowledge base and high-fidelity rendering capabilities. Gemini2.5-Flash-Image and SEEDream 3.0 also perform well. This track proves challenging for models with weaker world knowledge foundations, highlighting the importance of large-scale, high-quality training data for accurate real-world depictions.

\paragraph{Text Rendering}
Text rendering remains a significant challenge for almost all T2I models. Our benchmark confirms this, with this category receiving the lowest overall scores across all tracks. Notably, autoregressive models like Bagel and JanusPro perform poorly in this track, highlighting the inherent limitations of autoregressive architectures in text rendering tasks.

\paragraph{Style}
GPT-Image-1 demonstrates excellent performance in this track, earning a score of 93.1. Most modern models perform relatively well in this track, showing high fidelity to the requested styles. The high scores of these models indicate that the ability to capture stylistic essence is more mature compared to other tasks such as text rendering.

\paragraph{Affection}
Top models show extraordinary capability in capturing emotions and atmosphere. Gemini2.5-Flash-Image leads with an impressive 92.1 score, with GPT-Image-1 and Qwen-Image following closely behind. Notably, FLUX.1-dev achieves the highest aesthetic score in this category, indicating its generated images are particularly effective at visually conveying emotions, even with slightly lower prompt alignment.

\begin{figure}[p]
    \makebox[\linewidth]{
        \includegraphics[width=1\linewidth]{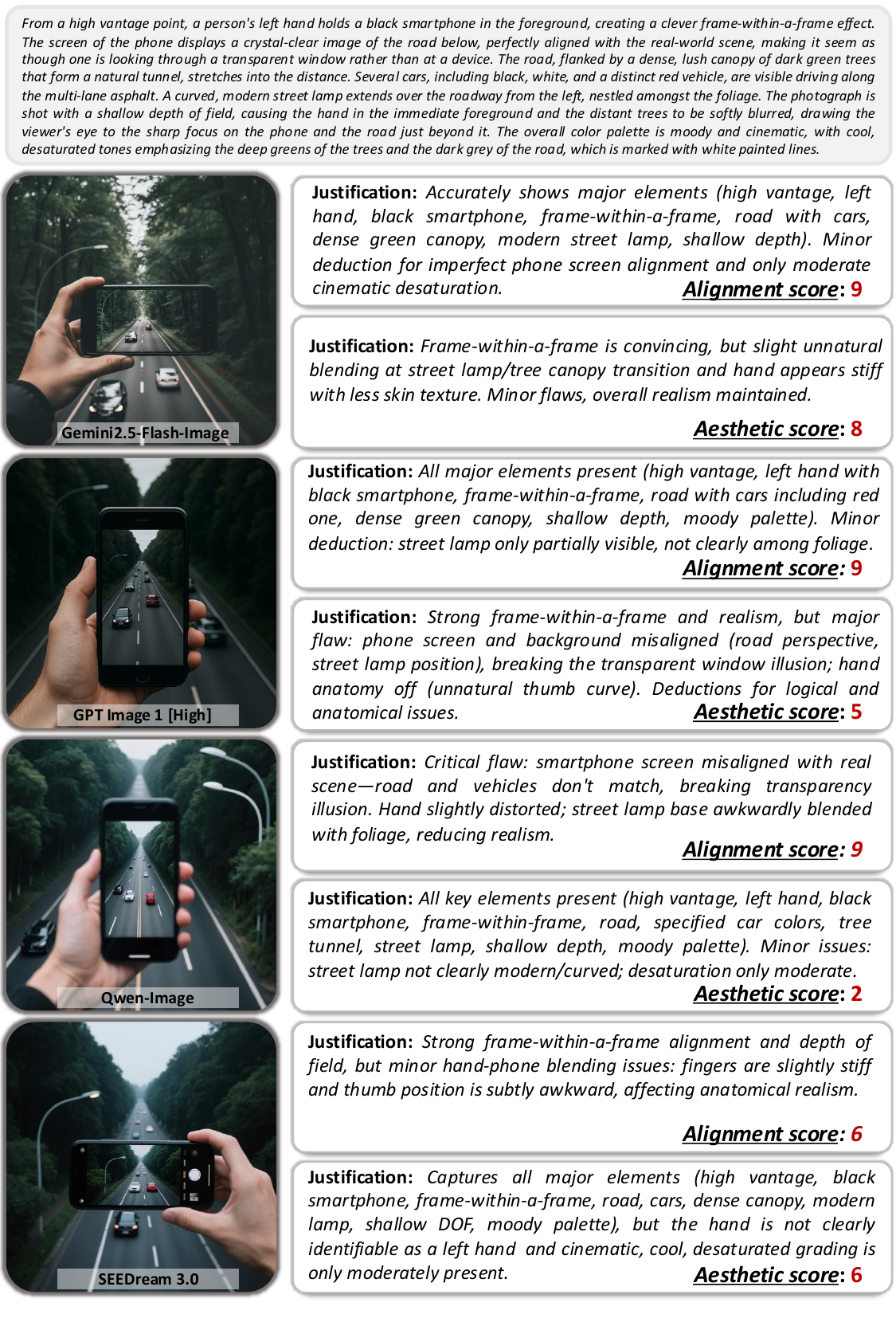}
    }
    \caption{Showcase of \textit{Long text} track in the PRISM-Bench. GPT4.1 is not only required to score based on image-text alignment and image aesthetics, but also to provide a brief justification.}
    \label{fig:result_gpt_en}
\end{figure}

\paragraph{Composition}
GPT-Image-1 leads by a wide margin with a high score of 92.8, fully demonstrating its ability to parse and execute complex spatial instructions. Gemini2.5-Flash-Image follows closely with a score of 90.5. Top open-source models are highly competitive in this domain. Qwen-Image scores almost identically to Gemini2.5-Flash-Image, indicating that the gap in complex compositional understanding is narrowing. Models like HiDream-I1-Full and FLUX.1-dev also demonstrate strong compositional capabilities. The small differences between the top performers suggest that composition control is becoming a mature capability in modern image generation systems.

\begin{table}[!t]\centering
\caption{Quantitative results on PRISM-Bench-ZH evaluated by GPT-4.1. The best result is in bold and the second best result is underlined.}
\renewcommand{\arraystretch}{1.7} 
\setlength{\tabcolsep}{3pt}

\centering
\begin{adjustbox}{width=\textwidth}
\begin{tabular}{l|ccc|ccc|ccc|ccc|ccc|ccc|ccc|ccc}
\toprule
\multirow{2}{*}{\textbf{Model}}
  & \multicolumn{3}{c|}{\textbf{Imagination}}
  & \multicolumn{3}{c|}{\textbf{Entity}}
  & \multicolumn{3}{c|}{\textbf{Text rendering}}
  & \multicolumn{3}{c|}{\textbf{Style}}
  & \multicolumn{3}{c|}{\textbf{Affection}}
  & \multicolumn{3}{c|}{\textbf{Composition}}
  & \multicolumn{3}{c|}{\textbf{Long text}}
  & \multicolumn{3}{c}{\textbf{Overall}} \\

\cmidrule(lr){2-25}

  & Ali. & Aes.& Avg.
  & Ali. & Aes.& Avg.   
  & Ali. & Aes.& Avg.     
  & Ali. & Aes.& Avg.     
  & Ali. & Aes.& Avg.     
  & Ali. & Aes.& Avg.     
  & Ali. & Aes.& Avg.     
  & Ali. & Aes.& Avg.     
  \\

\midrule

HiDream-I1-Dev~\citep{hidream} & 47.3 & 41.1 & 44.2 & 52.8 & 49.0 & 50.9 & 35.2 & 14.5 & 24.9 & 64.5 & 52.4 & 58.5 & 76.3 & 66.5 & 71.4 & 67.6 & 68.3 & 68.0 & 41.1 & 46.4 & 43.8 & 55.0 & 48.3 & 51.7 \\

HiDream-I1-Full~\citep{hidream} & 53.6 & 47.3 & 50.5 & 63.1 & 60.8 & 62.0 & 34.6 & 16.3 & 25.5 & 74.1 & 65.5 & 69.8 & 80.9 & 67.3 & 74.1 & 73.8 & 76.1 & 75.0 & 45.4 & 50.8 & 48.1 & 60.8 & 54.9 & 57.9 \\

Bagel-CoT~\citep{bagel} & 75.1 & 69.3 & 72.2 & 53.3 & 58.8 & 56.1 & 42.6 & 16.3 & 29.5 & 73.6 & 66.6 & 70.1 & 81.2 & 78.0 & 79.6 & 74.0 & 83.6 & 78.8 & 50.7 & 64.3 & 57.5 & 64.4 & 62.4 & 63.4 \\

Bagel~\citep{bagel} & 72.8 & 64.7 & 68.8 & 53.9 & 62.2 & 58.1 & 49.2 & 29.0 & 39.1 & 73.9 & 68.4 & 71.2 & 81.4 & 73.5 & 77.5 & 69.0 & 89.8 & 79.4 & 58.1 & 68.7 & 63.4 & 65.5 & 65.2 & 65.4 \\

Qwen-Image~\citep{QwenImage} & \underline{80.1} & \underline{79.6} & \underline{79.9} & 75.6 & \underline{79.7} & 77.7 & 76.9 & 62.9 & \underline{69.9} & \underline{90.2} & \underline{84.3} & \underline{87.3} & 87.4 & 84.9 & 86.2 & 86.6 & 93.4 & 90.0 & 68.9 & \underline{84.2} & 76.6 & 80.8 & 81.3 & 81.1 \\

SEEDream 3.0~\citep{seedream3} & 77.2 & 77.8 & 77.5 & \underline{77.6} & 78.6 & \underline{78.1} & \underline{79.7} & \textbf{71.9} & \textbf{75.8} & 87.8 & 83.2 & 85.5 & \underline{88.7} & \underline{85.1} & \underline{86.9} & \underline{87.7} & \underline{94.4} & \underline{91.1} & \underline{74.3} & 82.7 & \underline{78.5} & \underline{81.9} & \underline{82.0} & \underline{82.0} \\

GPT-Image-1 [High]~\citep{gptimage} & \textbf{88.8} & \textbf{90.4} & \textbf{89.6} & \textbf{85.9} & \textbf{92.4} & \textbf{89.2} & \textbf{83.9} & \underline{67.7} & \textbf{75.8} & \textbf{93.9} & \textbf{91.7} & \textbf{92.8} & \textbf{91.5} & \textbf{86.5} & \textbf{89.0} & \textbf{92.4} & \textbf{97.3} & \textbf{94.9} & \textbf{77.2} & \textbf{84.3} & \textbf{80.8} & \textbf{87.7} & \textbf{87.2} & \textbf{87.5} \\

\bottomrule

\end{tabular}\label{tab:gpt_zh}
\end{adjustbox}
\end{table}

\begin{table}[!t]\centering
\caption{Quantitative results on PRISM-Bench-ZH evaluated by Qwen2.5-VL-72B. The best result is in bold and the second best result is underlined.}
\renewcommand{\arraystretch}{1.7} 
\setlength{\tabcolsep}{3pt}

\centering
\begin{adjustbox}{width=\textwidth}
\begin{tabular}{l|ccc|ccc|ccc|ccc|ccc|ccc|ccc|ccc}
\toprule
\multirow{2}{*}{\textbf{Model}}
  & \multicolumn{3}{c|}{\textbf{Imagination}}
  & \multicolumn{3}{c|}{\textbf{Entity}}
  & \multicolumn{3}{c|}{\textbf{Text rendering}}
  & \multicolumn{3}{c|}{\textbf{Style}}
  & \multicolumn{3}{c|}{\textbf{Affection}}
  & \multicolumn{3}{c|}{\textbf{Composition}}
  & \multicolumn{3}{c|}{\textbf{Long text}}
  & \multicolumn{3}{c}{\textbf{Overall}} \\

\cmidrule(lr){2-25}

  & Ali. & Aes.& Avg.
  & Ali. & Aes.& Avg.   
  & Ali. & Aes.& Avg.     
  & Ali. & Aes.& Avg.     
  & Ali. & Aes.& Avg.     
  & Ali. & Aes.& Avg.     
  & Ali. & Aes.& Avg.     
  & Ali. & Aes.& Avg.     
  \\

\midrule

HiDream-I1-Dev~\citep{hidream} & 48.3 & 24.6 & 36.5 & 52.6 & 54.1 & 53.4 & 18.6 & 35.3 & 27.0 & 59.0 & 68.3 & 63.7 & 65.9 & 62.3 & 64.1 & 66.5 & 64.6 & 65.6 & 54.2 & 38.6 & 46.4 & 52.2 & 49.7 & 50.9 \\

HiDream-I1-Full~\citep{hidream} & 51.2 & 30.8 & 41.0 & 60.1 & 61.3 & 60.7 & 20.7 & 40.6 & 30.7 & 64.5 & 73.8 & 69.2 & 65.2 & 69.1 & 67.2 & 72.4 & 69.0 & 70.7 & 57.1 & 42.8 & 50.0 & 55.9 & 55.3 & 55.6 \\

Bagel~\citep{bagel} & 64.6 & 36.3 & 50.5 & 62.7 & 55.5 & 59.1 & 18.6 & 26.3 & 22.5 & 66.0 & 76.6 & 71.3 & 74.9 & 66.2 & 70.6 & 81.3 & 72.2 & 76.8 & 62.4 & 47.3 & 54.9 & 61.5 & 54.3 & 57.9 \\

Bagel-CoT~\citep{bagel} & 64.4 & \underline{36.6} & 50.5 & 62.6 & 53.8 & 58.2 & 25.2 & 51.9 & 38.6 & 65.4 & 76.7 & 71.1 & 74.0 & 65.0 & 69.5 & 81.3 & 71.3 & 76.3 & 61.4 & 46.6 & 54.0 & 62.0 & 57.4 & 59.7 \\

Qwen-Image~\citep{QwenImage} & \underline{71.4} & 29.9 & 50.7 & 74.7 & 67.8 & 71.3 & 64.3 & 73.1 & 68.7 & \underline{75.2} & 83.2 & 79.2 & 77.3 & 64.5 & 70.9 & 89.8 & 74.1 & 82.0 & \underline{72.6} & 65.8 & 69.2 & 75.0 & 65.5 & 70.3 \\

SEEDream 3.0~\citep{seedream3} & \underline{71.4} & \underline{36.6} & \underline{54.0} & \underline{74.8} & \underline{73.8} & \underline{74.3} & \underline{70.7} & \underline{88.0} & \underline{79.4} & 74.1 & \underline{88.0} & \underline{81.1} & \textbf{79.0} & \underline{71.4} & \underline{75.2} & \underline{90.30} & \underline{83.2} & \underline{86.8} & \textbf{73.0} & \underline{71.2} & \underline{72.1} & \underline{76.2} & \underline{73.2} & \underline{74.7} \\

GPT-Image-1 [High]~\citep{gptimage} & \textbf{73.0} & \textbf{37.6} & \textbf{55.3} & \textbf{80.4} & \textbf{82.1} & \textbf{81.3} & \textbf{73.1} & \textbf{89.9} & \textbf{81.5} & \textbf{77.1} & \textbf{92.4} & \textbf{84.8} & \underline{78.0} & \textbf{77.8} & \textbf{77.9} & \textbf{91.9} & \textbf{85.7} & \textbf{88.8} & 72.4 & \textbf{76.3} & \textbf{74.4} & \textbf{78.0} & \textbf{77.4} & \textbf{77.7} \\

\bottomrule

\end{tabular}\label{tab:qwen_zh}
\end{adjustbox}
\end{table}

\paragraph{Long Text}
The evaluation results clearly differentiate top models. Gemini2.5-Flash-Image achieves the highest score of 81.1, with GPT-Image-1 and SEEDream 3.0 also performing relatively well. However, compared to other tracks, the overall scores of all models are notably lower, indicating significant room for improvement in the ability to produce high-quality images following complex, multi-layered instructions in prompts. A typical example is presented in Figure~\ref{fig:result_gpt_en}. This highlights the reasoning gap issue that FLUX-Reason-6M aims to address.

\begin{figure}[p]
    \makebox[\linewidth]{
        \includegraphics[width=1\linewidth]{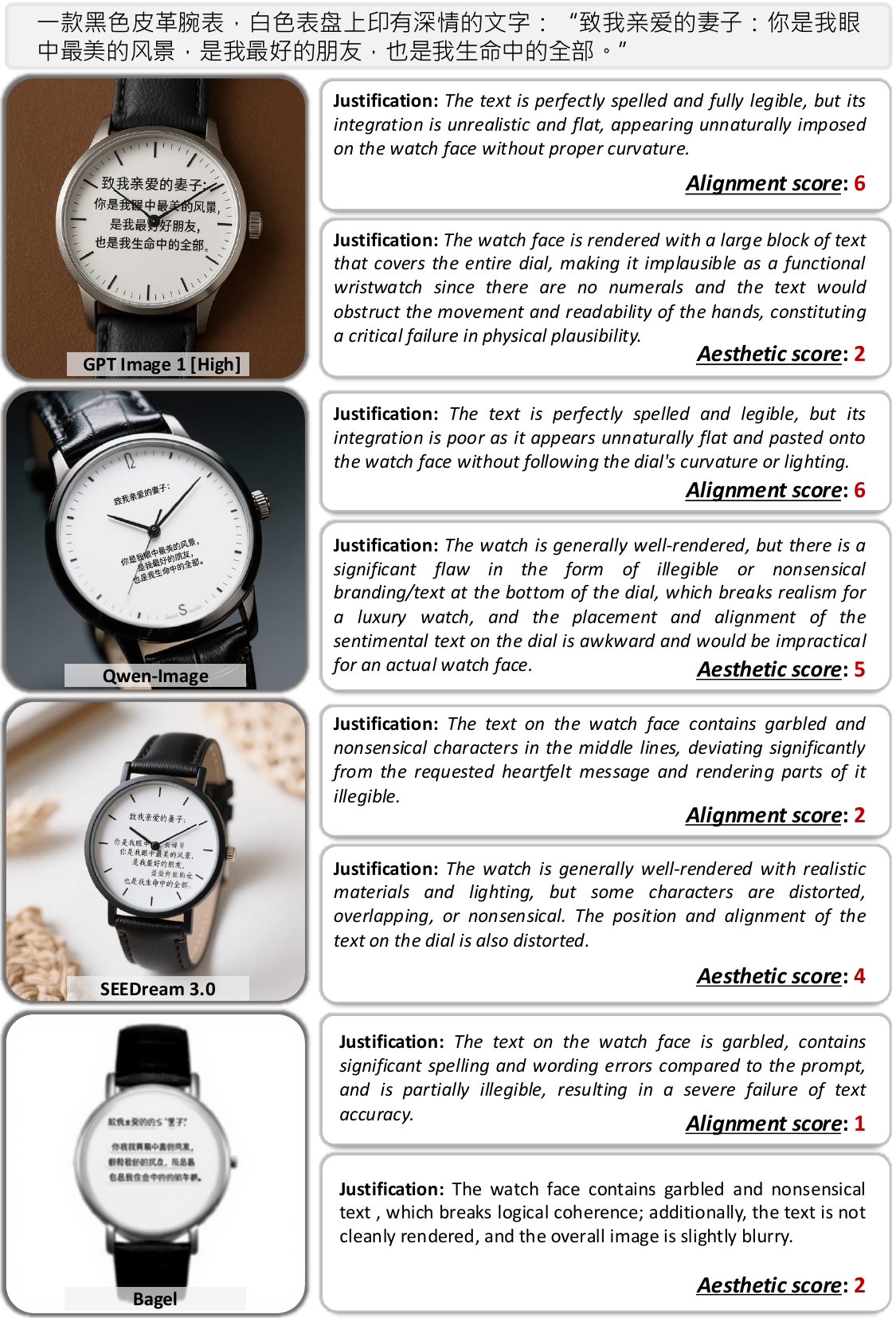}
    }
    \caption{Showcase of \textit{Text rendering} track in the PRISM-Bench-ZH.
     }
    \label{fig:result_gpt_zh}
\end{figure}

\subsection{Results and Analysis on PRISM-Bench-ZH}
The evaluation results from PRISM-Bench-ZH reveal a distinct performance hierarchy, with GPT-Image-1 establishing its dominance at a total score of 87.5. It consistently leads across most tracks, including \textit{Imagination}, \textit{Entity}, \textit{Style}, \textit{Affection}, and \textit{Composition}, demonstrating exceptional creative interpretation, knowledge foundation, and spatial arrangement in response to Chinese prompts. Meanwhile, SEEDream 3.0 and Qwen-Image demonstrate strong competitiveness across all tracks, frequently performing nearly on par with the leader. Particularly noteworthy is the performance of SEEDream 3.0 and Qwen-Image in \textit{Text rendering}, which stands in stark contrast to the general weakness observed in English text generation. Among these, SEEDream 3.0 and GPT-Image-1 share the highest average score, with SEEDream 3.0 achieving the highest aesthetic score, indicating its capability to render high-quality Chinese characters. The robust performance of these models validates the benchmark design's use of culturally adaptive prompts in Chinese and highlights significant advancements in handling Chinese typography. We present examples of Chinese text rendering across different models in Figure~\ref{fig:result_gpt_zh}. Nevertheless, consistent with PRISM-Bench test results, the \textit{Long text} track remains the greatest challenge for all models. While GPT-Image-1 again leads in this category, the generally lower scores highlight the substantial obstacle of understanding and synthesizing lengthy, multifaceted Chinese instructions. This further emphasizes the urgent need for reasoning-focused datasets like FLUX-Reason-6M to address existing gaps and train the next generation of truly intelligent T2I models.
 
\section{Conclusion}
In this work, we address critical gaps in text-to-image models through two key contributions: the FLUX-Reason-6M dataset and the PRISM benchmark. FLUX-Reason-6M is an extensive 6-million-image dataset with 20 million high-quality prompts specifically designed for reasoning, featuring novel generation chain-of-thought that imparts image synthesis logic across six characteristics to models. To measure progress, we develop PRISM-Bench, a comprehensive seven-track benchmark utilizing advanced VLMs for fine-grained human-aligned evaluation. Our extensive experimentation across 19 models reveals that while leading closed-source systems demonstrate impressive performance, all models struggle with complex tasks such as text rendering and long instruction following, underscoring the necessity of our work. By publicly releasing the dataset, benchmark, and evaluation code, we provide the community with essential tools for training and evaluating the next generation of more intelligent and capable T2I models.

\clearpage
\bibliography{colm2024_conference}

\begin{thebibliography}{72}
\providecommand{\natexlab}[1]{#1}
\providecommand{\url}[1]{\texttt{#1}}
\expandafter\ifx\csname urlstyle\endcsname\relax
  \providecommand{\doi}[1]{doi: #1}\else
  \providecommand{\doi}{doi: \begingroup \urlstyle{rm}\Url}\fi

\bibitem[Bai et~al.(2025)Bai, Chen, Liu, Wang, Ge, Song, Dang, Wang, Wang, Tang, et~al.]{Qwen2.5-VL}
Shuai Bai, Keqin Chen, Xuejing Liu, Jialin Wang, Wenbin Ge, Sibo Song, Kai Dang, Peng Wang, Shijie Wang, Jun Tang, et~al.
\newblock Qwen2. 5-vl technical report.
\newblock \emph{arXiv preprint arXiv:2502.13923}, 2025.

\bibitem[Bakr et~al.(2023)Bakr, Sun, Shen, Khan, Li, and Elhoseiny]{bakr2023hrs}
Eslam~Mohamed Bakr, Pengzhan Sun, Xiaoqian Shen, Faizan~Farooq Khan, Li~Erran Li, and Mohamed Elhoseiny.
\newblock Hrs-bench: Holistic, reliable and scalable benchmark for text-to-image models.
\newblock In \emph{Proceedings of the IEEE/CVF International Conference on Computer Vision}, pp.\  20041--20053, 2023.

\bibitem[BlackForest(2024)]{flux}
BlackForest.
\newblock Flux, 2024.
\newblock URL \url{https://github.com/black-forest-labs/flux}.

\bibitem[BlackForest(2025)]{fluxkrea}
BlackForest.
\newblock Flux.1 krea, 2025.
\newblock URL \url{https://www.krea.ai/apps/image/flux-krea}.

\bibitem[Cai et~al.(2025)Cai, Chen, Chen, Li, Long, Pan, Qiu, Zhang, Gao, Xu, et~al.]{hidream}
Qi~Cai, Jingwen Chen, Yang Chen, Yehao Li, Fuchen Long, Yingwei Pan, Zhaofan Qiu, Yiheng Zhang, Fengbin Gao, Peihan Xu, et~al.
\newblock Hidream-i1: A high-efficient image generative foundation model with sparse diffusion transformer.
\newblock \emph{arXiv preprint arXiv:2505.22705}, 2025.

\bibitem[Changpinyo et~al.(2021)Changpinyo, Sharma, Ding, and Soricut]{changpinyo2021conceptual}
Soravit Changpinyo, Piyush Sharma, Nan Ding, and Radu Soricut.
\newblock Conceptual 12m: Pushing web-scale image-text pre-training to recognize long-tail visual concepts.
\newblock In \emph{Proceedings of the IEEE/CVF conference on computer vision and pattern recognition}, pp.\  3558--3568, 2021.

\bibitem[Chefer et~al.(2023)Chefer, Alaluf, Vinker, Wolf, and Cohen-Or]{chefer2023attend}
Hila Chefer, Yuval Alaluf, Yael Vinker, Lior Wolf, and Daniel Cohen-Or.
\newblock Attend-and-excite: Attention-based semantic guidance for text-to-image diffusion models.
\newblock \emph{ACM transactions on Graphics (TOG)}, 42\penalty0 (4):\penalty0 1--10, 2023.

\bibitem[Chen et~al.(2023{\natexlab{a}})Chen, Huang, Lv, Cui, Chen, and Wei]{chen2023textdiffuser}
Jingye Chen, Yupan Huang, Tengchao Lv, Lei Cui, Qifeng Chen, and Furu Wei.
\newblock Textdiffuser: Diffusion models as text painters.
\newblock \emph{Advances in Neural Information Processing Systems}, 36:\penalty0 9353--9387, 2023{\natexlab{a}}.

\bibitem[Chen et~al.(2025{\natexlab{a}})Chen, Xu, Pan, Hu, Qin, Goldstein, Huang, Zhou, Xie, Savarese, et~al.]{chen2025blip3}
Jiuhai Chen, Zhiyang Xu, Xichen Pan, Yushi Hu, Can Qin, Tom Goldstein, Lifu Huang, Tianyi Zhou, Saining Xie, Silvio Savarese, et~al.
\newblock Blip3-o: A family of fully open unified multimodal models-architecture, training and dataset.
\newblock \emph{arXiv preprint arXiv:2505.09568}, 2025{\natexlab{a}}.

\bibitem[Chen et~al.(2023{\natexlab{b}})Chen, Yu, Ge, Yao, Xie, Wu, Wang, Kwok, Luo, Lu, et~al.]{chen2023pixart}
Junsong Chen, Jincheng Yu, Chongjian Ge, Lewei Yao, Enze Xie, Yue Wu, Zhongdao Wang, James Kwok, Ping Luo, Huchuan Lu, et~al.
\newblock Pixart-alpha: Fast training of diffusion transformer for photorealistic text-to-image synthesis.
\newblock \emph{arXiv preprint arXiv:2310.00426}, 2023{\natexlab{b}}.

\bibitem[Chen et~al.(2025{\natexlab{b}})Chen, Lai, Gao, Ye, Chen, Shi, Shao, Lin, Fei, Xing, et~al.]{chen2025postercraft}
SiXiang Chen, Jianyu Lai, Jialin Gao, Tian Ye, Haoyu Chen, Hengyu Shi, Shitong Shao, Yunlong Lin, Song Fei, Zhaohu Xing, et~al.
\newblock Postercraft: Rethinking high-quality aesthetic poster generation in a unified framework.
\newblock \emph{arXiv preprint arXiv:2506.10741}, 2025{\natexlab{b}}.

\bibitem[Chen et~al.(2025{\natexlab{c}})Chen, Wu, Liu, Pan, Liu, Xie, Yu, and Ruan]{januspro}
Xiaokang Chen, Zhiyu Wu, Xingchao Liu, Zizheng Pan, Wen Liu, Zhenda Xie, Xingkai Yu, and Chong Ruan.
\newblock Janus-pro: Unified multimodal understanding and generation with data and model scaling.
\newblock \emph{arXiv preprint arXiv:2501.17811}, 2025{\natexlab{c}}.

\bibitem[Cho et~al.(2023)Cho, Hu, Garg, Anderson, Krishna, Baldridge, Bansal, Pont-Tuset, and Wang]{cho2023davidsonian}
Jaemin Cho, Yushi Hu, Roopal Garg, Peter Anderson, Ranjay Krishna, Jason Baldridge, Mohit Bansal, Jordi Pont-Tuset, and Su~Wang.
\newblock Davidsonian scene graph: Improving reliability in fine-grained evaluation for text-to-image generation.
\newblock \emph{arXiv preprint arXiv:2310.18235}, 2023.

\bibitem[Deng et~al.(2025)Deng, Zhu, Li, Gou, Li, Wang, Zhong, Yu, Nie, Song, et~al.]{bagel}
Chaorui Deng, Deyao Zhu, Kunchang Li, Chenhui Gou, Feng Li, Zeyu Wang, Shu Zhong, Weihao Yu, Xiaonan Nie, Ziang Song, et~al.
\newblock Emerging properties in unified multimodal pretraining.
\newblock \emph{arXiv preprint arXiv:2505.14683}, 2025.

\bibitem[Duan et~al.(2025)Duan, Fang, Wang, Wang, Huang, Zeng, Li, and Liu]{duan2025got}
Chengqi Duan, Rongyao Fang, Yuqing Wang, Kun Wang, Linjiang Huang, Xingyu Zeng, Hongsheng Li, and Xihui Liu.
\newblock Got-r1: Unleashing reasoning capability of mllm for visual generation with reinforcement learning.
\newblock \emph{arXiv preprint arXiv:2505.17022}, 2025.

\bibitem[Esser et~al.(2024)Esser, Kulal, Blattmann, Entezari, M{\"u}ller, Saini, Levi, Lorenz, Sauer, Boesel, et~al.]{esser2024scaling}
Patrick Esser, Sumith Kulal, Andreas Blattmann, Rahim Entezari, Jonas M{\"u}ller, Harry Saini, Yam Levi, Dominik Lorenz, Axel Sauer, Frederic Boesel, et~al.
\newblock Scaling rectified flow transformers for high-resolution image synthesis.
\newblock In \emph{Forty-first international conference on machine learning}, 2024.

\bibitem[Fang et~al.(2024)Fang, Duan, Wang, Li, Tian, Zeng, Zhao, Dai, Li, and Liu]{fang2024puma}
Rongyao Fang, Chengqi Duan, Kun Wang, Hao Li, Hao Tian, Xingyu Zeng, Rui Zhao, Jifeng Dai, Hongsheng Li, and Xihui Liu.
\newblock Puma: Empowering unified mllm with multi-granular visual generation.
\newblock \emph{arXiv preprint arXiv:2410.13861}, 2024.

\bibitem[Fang et~al.(2025)Fang, Duan, Wang, Huang, Li, Yan, Tian, Zeng, Zhao, Dai, et~al.]{fang2025got}
Rongyao Fang, Chengqi Duan, Kun Wang, Linjiang Huang, Hao Li, Shilin Yan, Hao Tian, Xingyu Zeng, Rui Zhao, Jifeng Dai, et~al.
\newblock Got: Unleashing reasoning capability of multimodal large language model for visual generation and editing.
\newblock \emph{arXiv preprint arXiv:2503.10639}, 2025.

\bibitem[Feng et~al.(2022)Feng, He, Fu, Jampani, Akula, Narayana, Basu, Wang, and Wang]{feng2022training}
Weixi Feng, Xuehai He, Tsu-Jui Fu, Varun Jampani, Arjun Akula, Pradyumna Narayana, Sugato Basu, Xin~Eric Wang, and William~Yang Wang.
\newblock Training-free structured diffusion guidance for compositional text-to-image synthesis.
\newblock \emph{arXiv preprint arXiv:2212.05032}, 2022.

\bibitem[Fu et~al.(2024)Fu, He, Lu, Wang, and Roth]{fu2024commonsense}
Xingyu Fu, Muyu He, Yujie Lu, William~Yang Wang, and Dan Roth.
\newblock Commonsense-t2i challenge: Can text-to-image generation models understand commonsense?
\newblock \emph{arXiv preprint arXiv:2406.07546}, 2024.

\bibitem[Gadre et~al.(2023)Gadre, Ilharco, Fang, Hayase, Smyrnis, Nguyen, Marten, Wortsman, Ghosh, Zhang, et~al.]{gadre2023datacomp}
Samir~Yitzhak Gadre, Gabriel Ilharco, Alex Fang, Jonathan Hayase, Georgios Smyrnis, Thao Nguyen, Ryan Marten, Mitchell Wortsman, Dhruba Ghosh, Jieyu Zhang, et~al.
\newblock Datacomp: In search of the next generation of multimodal datasets.
\newblock \emph{Advances in Neural Information Processing Systems}, 36:\penalty0 27092--27112, 2023.

\bibitem[Gao et~al.(2025)Gao, Gong, Guo, Hou, Lai, Li, Li, Lian, Liao, Liu, et~al.]{seedream3}
Yu~Gao, Lixue Gong, Qiushan Guo, Xiaoxia Hou, Zhichao Lai, Fanshi Li, Liang Li, Xiaochen Lian, Chao Liao, Liyang Liu, et~al.
\newblock Seedream 3.0 technical report.
\newblock \emph{arXiv preprint arXiv:2504.11346}, 2025.

\bibitem[Ghosh et~al.(2023)Ghosh, Hajishirzi, and Schmidt]{ghosh2023geneval}
Dhruba Ghosh, Hannaneh Hajishirzi, and Ludwig Schmidt.
\newblock Geneval: An object-focused framework for evaluating text-to-image alignment.
\newblock \emph{Advances in Neural Information Processing Systems}, 36:\penalty0 52132--52152, 2023.

\bibitem[Gong et~al.(2025)Gong, Hou, Li, Li, Lian, Liu, Liu, Liu, Lu, Shi, et~al.]{gong2025seedream}
Lixue Gong, Xiaoxia Hou, Fanshi Li, Liang Li, Xiaochen Lian, Fei Liu, Liyang Liu, Wei Liu, Wei Lu, Yichun Shi, et~al.
\newblock Seedream 2.0: A native chinese-english bilingual image generation foundation model.
\newblock \emph{arXiv preprint arXiv:2503.07703}, 2025.

\bibitem[Google(2025{\natexlab{a}})]{gemini25}
Google.
\newblock Gemini2.5-pro, 2025{\natexlab{a}}.
\newblock URL \url{https://deepmind.google/models/gemini/pro/}.

\bibitem[Google(2025{\natexlab{b}})]{imagen4}
Google.
\newblock Imagen4, 2025{\natexlab{b}}.
\newblock URL \url{https://deepmind.google/models/imagen/}.

\bibitem[Google(2025{\natexlab{c}})]{nanobanana}
Google.
\newblock Gemini2.5-flash-image, 2025{\natexlab{c}}.
\newblock URL \url{https://deepmind.google/models/gemini/image/}.

\bibitem[Han et~al.(2024)Han, Fan, Fu, Li, Li, Cui, Wang, Tai, Sun, Guo, et~al.]{han2024evalmuse}
Shuhao Han, Haotian Fan, Jiachen Fu, Liang Li, Tao Li, Junhui Cui, Yunqiu Wang, Yang Tai, Jingwei Sun, Chunle Guo, et~al.
\newblock Evalmuse-40k: A reliable and fine-grained benchmark with comprehensive human annotations for text-to-image generation model evaluation.
\newblock \emph{arXiv preprint arXiv:2412.18150}, 2024.

\bibitem[Hessel et~al.(2021)Hessel, Holtzman, Forbes, Bras, and Choi]{hessel2021clipscore}
Jack Hessel, Ari Holtzman, Maxwell Forbes, Ronan~Le Bras, and Yejin Choi.
\newblock Clipscore: A reference-free evaluation metric for image captioning.
\newblock \emph{arXiv preprint arXiv:2104.08718}, 2021.

\bibitem[Ho et~al.(2020)Ho, Jain, and Abbeel]{ho2020denoising}
Jonathan Ho, Ajay Jain, and Pieter Abbeel.
\newblock Denoising diffusion probabilistic models.
\newblock \emph{Advances in neural information processing systems}, 33:\penalty0 6840--6851, 2020.

\bibitem[Hu et~al.(2022)Hu, Gan, Wang, Yang, Liu, Lu, and Wang]{hu2022scaling}
Xiaowei Hu, Zhe Gan, Jianfeng Wang, Zhengyuan Yang, Zicheng Liu, Yumao Lu, and Lijuan Wang.
\newblock Scaling up vision-language pre-training for image captioning.
\newblock In \emph{Proceedings of the IEEE/CVF conference on computer vision and pattern recognition}, pp.\  17980--17989, 2022.

\bibitem[Hu et~al.(2024)Hu, Wang, Fang, Fu, Cheng, and Yu]{hu2024ella}
Xiwei Hu, Rui Wang, Yixiao Fang, Bin Fu, Pei Cheng, and Gang Yu.
\newblock Ella: Equip diffusion models with llm for enhanced semantic alignment.
\newblock \emph{arXiv preprint arXiv:2403.05135}, 2024.

\bibitem[Hu et~al.(2023)Hu, Liu, Kasai, Wang, Ostendorf, Krishna, and Smith]{hu2023tifa}
Yushi Hu, Benlin Liu, Jungo Kasai, Yizhong Wang, Mari Ostendorf, Ranjay Krishna, and Noah~A Smith.
\newblock Tifa: Accurate and interpretable text-to-image faithfulness evaluation with question answering.
\newblock In \emph{Proceedings of the IEEE/CVF International Conference on Computer Vision}, pp.\  20406--20417, 2023.

\bibitem[Huang et~al.(2023)Huang, Sun, Xie, Li, and Liu]{huang2023t2i}
Kaiyi Huang, Kaiyue Sun, Enze Xie, Zhenguo Li, and Xihui Liu.
\newblock T2i-compbench: A comprehensive benchmark for open-world compositional text-to-image generation.
\newblock \emph{Advances in Neural Information Processing Systems}, 36:\penalty0 78723--78747, 2023.

\bibitem[Jia et~al.(2021)Jia, Yang, Xia, Chen, Parekh, Pham, Le, Sung, Li, and Duerig]{jia2021scaling}
Chao Jia, Yinfei Yang, Ye~Xia, Yi-Ting Chen, Zarana Parekh, Hieu Pham, Quoc Le, Yun-Hsuan Sung, Zhen Li, and Tom Duerig.
\newblock Scaling up visual and vision-language representation learning with noisy text supervision.
\newblock In \emph{International conference on machine learning}, pp.\  4904--4916. PMLR, 2021.

\bibitem[Krishna \& Murty(1999)Krishna and Murty]{krishna1999genetic}
KMMN Krishna and M~Narasimha Murty.
\newblock Genetic k-means algorithm.
\newblock \emph{IEEE Transactions on Systems, Man, and Cybernetics, Part B (Cybernetics)}, 29\penalty0 (3):\penalty0 433--439, 1999.

\bibitem[Li et~al.(2024{\natexlab{a}})Li, Lin, Pathak, Li, Fei, Wu, Ling, Xia, Zhang, Neubig, et~al.]{li2024genai}
Baiqi Li, Zhiqiu Lin, Deepak Pathak, Jiayao Li, Yixin Fei, Kewen Wu, Tiffany Ling, Xide Xia, Pengchuan Zhang, Graham Neubig, et~al.
\newblock Genai-bench: Evaluating and improving compositional text-to-visual generation.
\newblock \emph{arXiv preprint arXiv:2406.13743}, 2024{\natexlab{a}}.

\bibitem[Li et~al.(2024{\natexlab{b}})Li, Kamko, Akhgari, Sabet, Xu, and Doshi]{li2024playground}
Daiqing Li, Aleks Kamko, Ehsan Akhgari, Ali Sabet, Linmiao Xu, and Suhail Doshi.
\newblock Playground v2. 5: Three insights towards enhancing aesthetic quality in text-to-image generation.
\newblock \emph{arXiv preprint arXiv:2402.17245}, 2024{\natexlab{b}}.

\bibitem[Li et~al.(2020)Li, Yin, Li, Zhang, Hu, Zhang, Wang, Hu, Dong, Wei, et~al.]{li2020oscar}
Xiujun Li, Xi~Yin, Chunyuan Li, Pengchuan Zhang, Xiaowei Hu, Lei Zhang, Lijuan Wang, Houdong Hu, Li~Dong, Furu Wei, et~al.
\newblock Oscar: Object-semantics aligned pre-training for vision-language tasks.
\newblock In \emph{European conference on computer vision}, pp.\  121--137. Springer, 2020.

\bibitem[Li et~al.(2024{\natexlab{c}})Li, Zhang, Lin, Xiong, Long, Deng, Zhang, Liu, Huang, Xiao, et~al.]{li2024hunyuan}
Zhimin Li, Jianwei Zhang, Qin Lin, Jiangfeng Xiong, Yanxin Long, Xinchi Deng, Yingfang Zhang, Xingchao Liu, Minbin Huang, Zedong Xiao, et~al.
\newblock Hunyuan-dit: A powerful multi-resolution diffusion transformer with fine-grained chinese understanding.
\newblock \emph{arXiv preprint arXiv:2405.08748}, 2024{\natexlab{c}}.

\bibitem[Liang et~al.(2022)Liang, Wu, Hu, Gan, Wang, Wang, Liu, Fang, and Duan]{liang2022nuwa}
Jian Liang, Chenfei Wu, Xiaowei Hu, Zhe Gan, Jianfeng Wang, Lijuan Wang, Zicheng Liu, Yuejian Fang, and Nan Duan.
\newblock Nuwa-infinity: Autoregressive over autoregressive generation for infinite visual synthesis.
\newblock \emph{Advances in Neural Information Processing Systems}, 35:\penalty0 15420--15432, 2022.

\bibitem[Lin et~al.(2024)Lin, Pathak, Li, Li, Xia, Neubig, Zhang, and Ramanan]{lin2024evaluating}
Zhiqiu Lin, Deepak Pathak, Baiqi Li, Jiayao Li, Xide Xia, Graham Neubig, Pengchuan Zhang, and Deva Ramanan.
\newblock Evaluating text-to-visual generation with image-to-text generation.
\newblock In \emph{European Conference on Computer Vision}, pp.\  366--384. Springer, 2024.

\bibitem[Liu et~al.(2022)Liu, Gong, and Liu]{liu2022flow}
Xingchao Liu, Chengyue Gong, and Qiang Liu.
\newblock Flow straight and fast: Learning to generate and transfer data with rectified flow.
\newblock \emph{arXiv preprint arXiv:2209.03003}, 2022.

\bibitem[Mohsin \& Beltiukov(2019)Mohsin and Beltiukov]{mohsin2019summarizing}
Manshad~Abbasi Mohsin and Anatoly Beltiukov.
\newblock Summarizing emotions from text using plutchik’s wheel of emotions.
\newblock In \emph{7th scientific conference on information technologies for intelligent decision making support (ITIDS 2019)}, pp.\  291--294. Atlantis Press, 2019.

\bibitem[OpenAI(2025{\natexlab{a}})]{gpt4-1}
OpenAI.
\newblock Gpt-4.1, 2025{\natexlab{a}}.
\newblock URL \url{https://openai.com/index/gpt-4-1/}.

\bibitem[OpenAI(2025{\natexlab{b}})]{gptimage}
OpenAI.
\newblock Gpt-image-1, 2025{\natexlab{b}}.
\newblock URL \url{https://openai.com/index/introducing-4o-image-generation/}.

\bibitem[OpenAI(September 2023)]{dalle3}
OpenAI.
\newblock Dall·e 3, September 2023.
\newblock URL \url{https://openai.com/zh-Hans-CN/index/dall-e-3/}.

\bibitem[Podell et~al.(2023)Podell, English, Lacey, Blattmann, Dockhorn, M{\"u}ller, Penna, and Rombach]{podell2023sdxl}
Dustin Podell, Zion English, Kyle Lacey, Andreas Blattmann, Tim Dockhorn, Jonas M{\"u}ller, Joe Penna, and Robin Rombach.
\newblock Sdxl: Improving latent diffusion models for high-resolution image synthesis.
\newblock \emph{arXiv preprint arXiv:2307.01952}, 2023.

\bibitem[Qin et~al.(2025)Qin, Zhuo, Xin, Du, Li, Fu, Lu, Yuan, Li, Liu, et~al.]{qin2025lumina}
Qi~Qin, Le~Zhuo, Yi~Xin, Ruoyi Du, Zhen Li, Bin Fu, Yiting Lu, Jiakang Yuan, Xinyue Li, Dongyang Liu, et~al.
\newblock Lumina-image 2.0: A unified and efficient image generative framework.
\newblock \emph{arXiv preprint arXiv:2503.21758}, 2025.

\bibitem[Rombach et~al.(2022)Rombach, Blattmann, Lorenz, Esser, and Ommer]{rombach2022high}
Robin Rombach, Andreas Blattmann, Dominik Lorenz, Patrick Esser, and Bj{\"o}rn Ommer.
\newblock High-resolution image synthesis with latent diffusion models.
\newblock In \emph{Proceedings of the IEEE/CVF conference on computer vision and pattern recognition}, pp.\  10684--10695, 2022.

\bibitem[Schuhmann et~al.(2022)Schuhmann, Beaumont, Vencu, Gordon, Wightman, Cherti, Coombes, Katta, Mullis, Wortsman, et~al.]{schuhmann2022laion}
Christoph Schuhmann, Romain Beaumont, Richard Vencu, Cade Gordon, Ross Wightman, Mehdi Cherti, Theo Coombes, Aarush Katta, Clayton Mullis, Mitchell Wortsman, et~al.
\newblock Laion-5b: An open large-scale dataset for training next generation image-text models.
\newblock \emph{Advances in neural information processing systems}, 35:\penalty0 25278--25294, 2022.

\bibitem[Sharma et~al.(2018)Sharma, Ding, Goodman, and Soricut]{sharma2018conceptual}
Piyush Sharma, Nan Ding, Sebastian Goodman, and Radu Soricut.
\newblock Conceptual captions: A cleaned, hypernymed, image alt-text dataset for automatic image captioning.
\newblock In \emph{Proceedings of the 56th Annual Meeting of the Association for Computational Linguistics (Volume 1: Long Papers)}, pp.\  2556--2565, 2018.

\bibitem[Singla et~al.(2024)Singla, Yue, Paul, Shirkavand, Jayawardhana, Ganjdanesh, Huang, Bhatele, Somepalli, and Goldstein]{singla2024pixels}
Vasu Singla, Kaiyu Yue, Sukriti Paul, Reza Shirkavand, Mayuka Jayawardhana, Alireza Ganjdanesh, Heng Huang, Abhinav Bhatele, Gowthami Somepalli, and Tom Goldstein.
\newblock From pixels to prose: A large dataset of dense image captions.
\newblock \emph{arXiv preprint arXiv:2406.10328}, 2024.

\bibitem[Stability-AI(2022)]{SD21}
Stability-AI.
\newblock Stable diffusion 2.1, 2022.
\newblock URL \url{https://huggingface.co/stabilityai/stable-diffusion-2-1}.

\bibitem[Stability-AI(2024{\natexlab{a}})]{SD3}
Stability-AI.
\newblock Stable diffusion 3, 2024{\natexlab{a}}.
\newblock URL \url{https://huggingface.co/stabilityai/stable-diffusion-3-medium}.

\bibitem[Stability-AI(2024{\natexlab{b}})]{SD35}
Stability-AI.
\newblock Stable diffusion 3.5, 2024{\natexlab{b}}.
\newblock URL \url{https://github.com/Stability-AI/sd3.5}.

\bibitem[Sun et~al.(2025)Sun, Fang, Duan, Liu, and Liu]{sun2025t2i}
Kaiyue Sun, Rongyao Fang, Chengqi Duan, Xian Liu, and Xihui Liu.
\newblock T2i-reasonbench: Benchmarking reasoning-informed text-to-image generation.
\newblock \emph{arXiv preprint arXiv:2508.17472}, 2025.

\bibitem[Sun et~al.(2023)Sun, Pan, Ge, Li, Duan, Wu, Zhang, Zhou, Qin, Wang, et~al.]{sun2023journeydb}
Keqiang Sun, Junting Pan, Yuying Ge, Hao Li, Haodong Duan, Xiaoshi Wu, Renrui Zhang, Aojun Zhou, Zipeng Qin, Yi~Wang, et~al.
\newblock Journeydb: A benchmark for generative image understanding.
\newblock \emph{Advances in neural information processing systems}, 36:\penalty0 49659--49678, 2023.

\bibitem[Tuo et~al.(2023)Tuo, Xiang, He, Geng, and Xie]{tuo2023anytext}
Yuxiang Tuo, Wangmeng Xiang, Jun-Yan He, Yifeng Geng, and Xuansong Xie.
\newblock Anytext: Multilingual visual text generation and editing.
\newblock \emph{arXiv preprint arXiv:2311.03054}, 2023.

\bibitem[Wang et~al.(2025)Wang, Mao, Zhang, Han, Dong, Li, Lin, Yang, Qin, Zhang, et~al.]{wang2025textatlas5m}
Alex~Jinpeng Wang, Dongxing Mao, Jiawei Zhang, Weiming Han, Zhuobai Dong, Linjie Li, Yiqi Lin, Zhengyuan Yang, Libo Qin, Fuwei Zhang, et~al.
\newblock Textatlas5m: A large-scale dataset for dense text image generation.
\newblock \emph{arXiv preprint arXiv:2502.07870}, 2025.

\bibitem[Wu et~al.(2022)Wu, Liang, Ji, Yang, Fang, Jiang, and Duan]{wu2022nuwa}
Chenfei Wu, Jian Liang, Lei Ji, Fan Yang, Yuejian Fang, Daxin Jiang, and Nan Duan.
\newblock Nuwa: Visual synthesis pre-training for neural visual world creation.
\newblock In \emph{European conference on computer vision}, pp.\  720--736. Springer, 2022.

\bibitem[Wu et~al.(2025{\natexlab{a}})Wu, Li, Zhou, Lin, Gao, Yan, Yin, Bai, Xu, Chen, et~al.]{QwenImage}
Chenfei Wu, Jiahao Li, Jingren Zhou, Junyang Lin, Kaiyuan Gao, Kun Yan, Sheng-ming Yin, Shuai Bai, Xiao Xu, Yilei Chen, et~al.
\newblock Qwen-image technical report.
\newblock \emph{arXiv preprint arXiv:2508.02324}, 2025{\natexlab{a}}.

\bibitem[Wu et~al.(2025{\natexlab{b}})Wu, Zheng, Yan, Xiao, Luo, Wang, Li, Jiang, Liu, Zhou, et~al.]{wu2025omnigen2}
Chenyuan Wu, Pengfei Zheng, Ruiran Yan, Shitao Xiao, Xin Luo, Yueze Wang, Wanli Li, Xiyan Jiang, Yexin Liu, Junjie Zhou, et~al.
\newblock Omnigen2: Exploration to advanced multimodal generation.
\newblock \emph{arXiv preprint arXiv:2506.18871}, 2025{\natexlab{b}}.

\bibitem[Wu et~al.(2023)Wu, Hao, Sun, Chen, Zhu, Zhao, and Li]{wu2023human}
Xiaoshi Wu, Yiming Hao, Keqiang Sun, Yixiong Chen, Feng Zhu, Rui Zhao, and Hongsheng Li.
\newblock Human preference score v2: A solid benchmark for evaluating human preferences of text-to-image synthesis.
\newblock \emph{arXiv preprint arXiv:2306.09341}, 2023.

\bibitem[Wu et~al.(2024)Wu, Yu, Huang, Russakovsky, and Arora]{wu2024conceptmix}
Xindi Wu, Dingli Yu, Yangsibo Huang, Olga Russakovsky, and Sanjeev Arora.
\newblock Conceptmix: A compositional image generation benchmark with controllable difficulty.
\newblock \emph{Advances in Neural Information Processing Systems}, 37:\penalty0 86004--86047, 2024.

\bibitem[Xie et~al.(2025{\natexlab{a}})Xie, Chen, Zhao, Yu, Zhu, Wu, Lin, Zhang, Li, Chen, et~al.]{xie2025sana}
Enze Xie, Junsong Chen, Yuyang Zhao, Jincheng Yu, Ligeng Zhu, Chengyue Wu, Yujun Lin, Zhekai Zhang, Muyang Li, Junyu Chen, et~al.
\newblock Sana 1.5: Efficient scaling of training-time and inference-time compute in linear diffusion transformer.
\newblock \emph{arXiv preprint arXiv:2501.18427}, 2025{\natexlab{a}}.

\bibitem[Xie et~al.(2025{\natexlab{b}})Xie, Yang, and Shou]{xie2025show}
Jinheng Xie, Zhenheng Yang, and Mike~Zheng Shou.
\newblock Show-o2: Improved native unified multimodal models.
\newblock \emph{arXiv preprint arXiv:2506.15564}, 2025{\natexlab{b}}.

\bibitem[Yang et~al.(2025)Yang, Li, Yang, Zhang, Hui, Zheng, Yu, Gao, Huang, Lv, et~al.]{qwen3}
An~Yang, Anfeng Li, Baosong Yang, Beichen Zhang, Binyuan Hui, Bo~Zheng, Bowen Yu, Chang Gao, Chengen Huang, Chenxu Lv, et~al.
\newblock Qwen3 technical report.
\newblock \emph{arXiv preprint arXiv:2505.09388}, 2025.

\bibitem[Yu et~al.(2022)Yu, Xu, Koh, Luong, Baid, Wang, Vasudevan, Ku, Yang, Ayan, et~al.]{yu2022scaling}
Jiahui Yu, Yuanzhong Xu, Jing~Yu Koh, Thang Luong, Gunjan Baid, Zirui Wang, Vijay Vasudevan, Alexander Ku, Yinfei Yang, Burcu~Karagol Ayan, et~al.
\newblock Scaling autoregressive models for content-rich text-to-image generation.
\newblock \emph{arXiv preprint arXiv:2206.10789}, 2\penalty0 (3):\penalty0 5, 2022.

\bibitem[Yu et~al.(2024)Yu, Sun, Zhang, Cui, Zhang, Cao, Wang, and Liu]{yu2024capsfusion}
Qiying Yu, Quan Sun, Xiaosong Zhang, Yufeng Cui, Fan Zhang, Yue Cao, Xinlong Wang, and Jingjing Liu.
\newblock Capsfusion: Rethinking image-text data at scale.
\newblock In \emph{Proceedings of the IEEE/CVF Conference on Computer Vision and Pattern Recognition}, pp.\  14022--14032, 2024.

\bibitem[Zhang et~al.(2021)Zhang, Li, Hu, Yang, Zhang, Wang, Choi, and Gao]{zhang2021vinvl}
Pengchuan Zhang, Xiujun Li, Xiaowei Hu, Jianwei Yang, Lei Zhang, Lijuan Wang, Yejin Choi, and Jianfeng Gao.
\newblock Vinvl: Making visual representations matter in vision-language models.
\newblock \emph{arXiv preprint arXiv:2101.00529}, 1\penalty0 (6):\penalty0 8, 2021.

\bibitem[Zhuo et~al.(2024)Zhuo, Du, Xiao, Li, Liu, Huang, Liu, Zhu, Wang, Ma, et~al.]{zhuo2024lumina}
Le~Zhuo, Ruoyi Du, Han Xiao, Yangguang Li, Dongyang Liu, Rongjie Huang, Wenze Liu, Xiangyang Zhu, Fu-Yun Wang, Zhanyu Ma, et~al.
\newblock Lumina-next: Making lumina-t2x stronger and faster with next-dit.
\newblock \emph{Advances in Neural Information Processing Systems}, 37:\penalty0 131278--131315, 2024.

\end{thebibliography}
\bibliographystyle{colm2024_conference}

\end{document}